\documentclass{article}

\usepackage[main, final]{iaseai26}

\usepackage[utf8]{inputenc}
\usepackage[T1]{fontenc}
\usepackage{hyperref}
\usepackage{url}
\usepackage{amsfonts}
\usepackage{nicefrac}
\usepackage{microtype}

\usepackage[table,xcdraw]{xcolor}
\definecolor{tblbg}{HTML}{F3F3F3}
\definecolor{tblhd}{HTML}{D9D9D9}

\usepackage{graphicx}
\usepackage{tabularx}
\usepackage{booktabs}
\usepackage{multirow}
\usepackage{makecell}
\usepackage{array}
\usepackage{longtable}
\usepackage{colortbl}
\usepackage{caption}
\usepackage{subcaption}
\usepackage{float}
\usepackage{placeins}
\usepackage{amsmath}

\usepackage{siunitx}
\usepackage{enumitem}

\title{Exposing Blindspots: Cultural Bias Evaluation in Generative Image Models}

\author{\mdseries
\textbf{Huichan Seo}\textsuperscript{1*}\qquad
\textbf{Sieun Choi}\textsuperscript{2*\textdagger}\qquad
\textbf{Minki Hong}\textsuperscript{2*\textdagger}\qquad
\textbf{Yi Zhou}\textsuperscript{1}\\
\textbf{Junseo Kim}\textsuperscript{3\textdagger}\qquad
\textbf{Lukman Ismaila}\textsuperscript{4}\qquad
\textbf{Naome Etori}\textsuperscript{5}\qquad
\textbf{Mehul Agarwal}\textsuperscript{1}\\
\textbf{Zhixuan Liu}\textsuperscript{1}\qquad
\textbf{Jihie Kim}\textsuperscript{2}\qquad
\textbf{Jean Oh}\textsuperscript{1,6}
\\[0.75ex]
\textsuperscript{1}Carnegie Mellon University, Pittsburgh, United States\\
\textsuperscript{2}Dongguk University, Seoul, South Korea\\
\textsuperscript{3}Delft University of Technology, Delft, Netherlands\\
\textsuperscript{4}Johns Hopkins University, School of Medicine, Baltimore, United States\\
\textsuperscript{5}University of Minnesota--Twin Cities, Minneapolis, United States\\
\textsuperscript{6}Lavoro AI, Pittsburgh, United States\\[0.75ex]
\texttt{Corresponding author: Jean Oh (\textit{jeanoh@cmu.edu})}
}
 
\begin{document}
 
\maketitle
\vspace{-0.1in}
\begingroup
\renewcommand{\thefootnote}{\fnsymbol{footnote}}
\footnotetext[1]{Equal contribution.}
\footnotetext[2]{This paper is based on the work performed while the authors were visiting scholars at Carnegie Mellon University.}
\endgroup
\begin{abstract}
Generative image models produce striking visuals yet often misrepresent culture. Prior work has probed cultural dimensions primarily in text-to-image (T2I) systems, leaving image-to-image (I2I) editors largely underexamined. We close this gap with a unified, reproducible evaluation spanning six countries, an 8-category/36-subcategory schema, and era-aware prompts, auditing both T2I generation and I2I editing under a standardized, reproducible protocol that yields comparable model-level diagnostics.
Using open models with fixed configurations, we derive comparable diagnostics across countries, eras, and categories for both T2I and I2I. Our evaluation combines standard automatic measures, a culture-aware metric that integrates retrieval-augmented visual question answering (VQA) with curated knowledge, and expert human judgments collected on a web platform from country-native reviewers. To enable downstream analyses without re-running compute-intensive pipelines, we release the complete image corpus from both studies alongside prompts and settings.
Our study reveals three recurring findings. First, under country-agnostic prompts, models default to U.S.-like, modern-leaning depictions and flatten cross-country distinctions, reducing separability between culturally distinct neighbors despite fixed schema and era controls. Second, iterative I2I editing erodes cultural fidelity even when conventional metrics remain flat or improve; by contrast, expert ratings and our culture-aware metric both register this degradation. Third, I2I models tend to apply superficial cues (palette shifts, generic props) rather than context- and era-consistent changes, frequently retaining source identity for non-U.S. targets and drifting toward non-photorealistic styles; attribute-addition trials further expose weak text rendering and brittle handling of fine, culture-specific details. Taken together, these results indicate that culture-sensitive edits remain unreliable in current systems. By standardizing prompts, settings, metrics, and human evaluation protocols---and releasing all images and configurations---we offer a reproducible, culture-centered pipeline for diagnosing and tracking progress in generative image research. Project page: \href{https://seochan99.github.io/ECB/}{https://seochan99.github.io/ECB/}.\end{abstract}

\section{Introduction}
\label{sec:1}
Text-to-image (T2I) and image-to-image (I2I) generative models have advanced rapidly in photorealism and controllability~\cite{saharia2022photorealistic, esser2024scaling, sordo2025review}. Yet T2I models exhibit systematic cultural bias rooted in imbalanced, skewed training data, where some regions and communities are over-represented while others are under-represented~\cite{kandwal2024survey, wan2024survey}. This yields stereotyped or inaccurate portrayals of underrepresented groups, as shown in Figure~\ref{fig:intro-bias-row}. I2I editing is often used as a practical remedy by inserting or adjusting culture-specific elements~\cite{khanuja2024image}, but errors persist across regions and frequently resurface over multiple edits~\cite{liu2023cultural, kannen2024beyond}. As a result, the burden of cultural correction shifts to users. We therefore focus on the \emph{I2I editing loop}: a sequence of edits intended to align an image with a target culture, used to examine whether bias persists, attenuates, or reappears across iterations.

\begingroup
\captionsetup[figure]{skip=\baselineskip}
\begin{figure}[t]
    \centering
    \includegraphics[width=\linewidth]{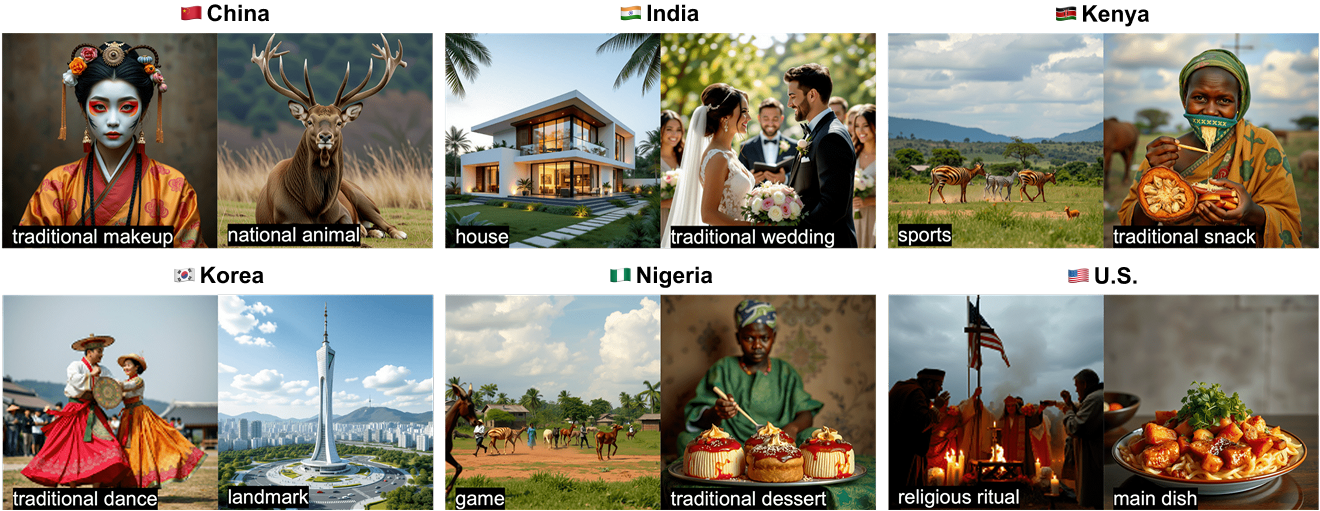}
    \caption{Representative cultural biases in T2I generations across six countries. Examples include Chinese–Japanese aesthetic conflation, mis-styled Indian weddings, Kenyan wildlife stereotypes, Korean attire misidentification, Nigerian safari mislocalization, and U.S. cultural miscues in food and religious ritual. Images are from FLUX.1 [schnell] fp8 and HiDream-I1-Dev.}
    \label{fig:intro-bias-row}
    \vspace{-8pt}
\end{figure}
\endgroup
\vspace{-3pt}

Turning to evaluation, cultural fidelity remains challenging to measure. Distributional or general-alignment metrics such as Fréchet Inception Distance (FID)~\cite{heusel2017gans} and CLIPScore~\cite{hessel2021clipscore} do not capture culture-specific attributes. Emerging cultural benchmarks~\cite{kannen2024beyond, bayramli2025diffusion} help, but often rely heavily on human studies or use only author-generated synthetic images, limiting generalization~\cite{nayak2025culturalframes}.

We analyze five representative model families (Table~\ref{tab:models_t2i_i2i}) by first constructing a T2I base image set spanning six countries: China, India, Kenya, Korea, Nigeria, and the United States (U.S.). The set is balanced across eight cultural categories (Architecture, Art, Events, Fashion, Food, Landscape, People, Wildlife), further stratified into 36 subcategories (Table~\ref{tab:category}). Our prompts are \emph{era-aware}, comprising \textit{traditional}, \textit{modern}, and \textit{era-agnostic} variants, which allows us to probe temporal sensitivity in cultural understanding. To the best of our knowledge, this is the first study to systematically evaluate era-aware cultural competence in image models. Building on the base image set, we evaluate I2I cultural adaptation using three complementary experiments: (1) Multi-Loop Edit: five sequential edits that test whether the model preserves context while progressively improving cultural consistency; (2) Attribute Addition: stepwise insertion of five distinct culture-specific attributes to quantify cumulative competence and interference; and (3) Cross-Country Restylization: coherent transfer from a source country to a target country to assess adaptability and generalization.

For evaluation, we adopt automatic metrics—CLIPScore~\cite{hessel2021clipscore}, DreamSim~\cite{fu2023dreamsim}, and Aesthetic Score~\cite{huang2024predicting}—complemented by culture-centered assessments. We compare automatic metrics against two complementary evaluations: a VQA-based culture-aware metric and a human evaluation. This approach allows us to analyze the agreement, gaps, and limitations between automatic evaluations and human perception.

Our controlled prompting method is designed to reduce variability from user phrasing and isolate model-side cultural tendencies, framing this study as a diagnostic stress test rather than a full model of all real-world usage.
We also conduct an additional bias analysis focusing on gender and skin-tone stereotyping that persists across generative models (see Appendix~\ref{app:G}). These findings point to a concrete need: curated and balanced training data, explicit training-time debiasing/regularization, and standardized fairness benchmarks to enable equitable and responsible deployment.

Our contributions are as follows:
\begin{enumerate}[itemsep=0pt, topsep=2pt, parsep=0pt, leftmargin=*]
  \item \textbf{Experimental design.} A geographically balanced, multi-domain schema (6 countries; 8 categories, 36 subcategories) and 3 complementary protocols (Multi-Loop Edit, Attribute Addition, Cross-Country Restylization) that contrast inherent T2I bias with I2I editing capability, including an era-aware prompt design to assess temporal awareness (an underexplored evaluation setting).
  \item \textbf{Dataset release.} Public release of the complete image set, together with prompts and model/execution configurations, enabling downstream cultural analyses without re-running generation/editing or model reconfiguration.
  \item \textbf{Evaluation framework.} An open-source evaluation platform that pairs automated metrics with culture-aware assessments and human studies, enabling triangulation across automated, culture-aware, and human judgments; we benchmark and validate metric behavior against human results to surface agreements and discrepancies.
\end{enumerate}

\section{Related Work}
\label{sec:2}

\subsection{Generative Image Models}
\label{sec:2.1}
Generative image models rely on two closely related paradigms: T2I generation and I2I editing. Latent diffusion models~\cite{rombach2022high, esser2024scaling} set the T2I standard for fidelity and controllability, yet often miss fine-grained cultural fidelity. In response, I2I frameworks increase edit precision via context-aware conditioning, multi-step guidance, and cross-attention manipulation (e.g., FLUX.1 Kontext~\cite{labs2025flux}, HiDream-I1~\cite{cai2025hidream}, Qwen-Image-Edit~\cite{wu2025qwen}, NextStep-1~\cite{team2025nextstep}). These methods preserve locality and enable targeted attributes but are typically optimized for realism and generic prompt compliance rather than culture-specific correctness. Prior studies~\cite{bianchi2023easily, naik2023social, chinchure2024tibet} show that stronger editing control alone does not mitigate entrenched biases: models can encode or even amplify stereotypes when data and objectives are not culture-aware. We therefore compare T2I baselines and their I2I editing loops to quantify initial cultural bias and the bias-correction burden placed on users.

\subsection{Text-Image Datasets}
\label{sec:2.2}
The foundation of modern generative image models rests on large-scale text-image datasets. Early efforts like MS-COCO~\cite{lin2014microsoft} and ImageNet~\cite{deng2009imagenet} were predominantly Western-centric. Subsequent expansions, including LAION-5B~\cite{schuhmann2022laion}, achieved unprecedented scale but inherently inherited severe cultural and geographical biases, notably English dominance and uneven global representation. While datasets like Dollar Street~\cite{gaviria2022dollar} and WIT~\cite{srinivasan2021wit} attempted to target diversity, their scope was limited by inherent editorial constraints. More recent benchmarks, including CCUB~\cite{liu2023towards} and SCoFT~\cite{liu2024scoft}, have focused on directly assessing cultural fidelity. However, these resources primarily prioritize concept coverage over the representation of nuanced cultural practices, ceremonies, or daily life contexts~\cite{nayak2024benchmarking}. This persistent gap---where existing datasets remain ill-suited for assessing authentic cultural representation---motivates the creation of our targeted evaluation framework, which is the direct focus of this work.

\subsection{Evaluation Metrics in Cultural Image Generation} 
\label{sec:2.3}
The evolution of generative image models mandates corresponding advancements in output evaluation. Traditional metrics mainly assess image quality (e.g., FID~\cite{heusel2017gans}, DreamSim~\cite{fu2023dreamsim}, Aesthetic Score~\cite{huang2024predicting}) and text-image alignment (e.g., CLIPScore~\cite{hessel2021clipscore} and DinoScore~\cite{ruiz2023dreambooth}). Recent improvements leverage human preference-trained models~\cite{wu2023human} and large language model (LLM)-based metrics~\cite{lin2024evaluating}. However, these approaches primarily capture realism and faithfulness to prompts, while neglecting cultural and contextual fidelity~\cite{kannen2024beyond, said2025deconstructing}. Recent benchmarks such as CUBE~\cite{kannen2024beyond}, CULTDIFF~\cite{bayramli2025diffusion}, and CulturalFrames~\cite{nayak2024benchmarking} aim to evaluate cultural alignment and bias, yet often depend on human judgments or limited internal datasets, constraining generalizability. To overcome this gap, we propose a comprehensive framework for evaluating authentic cultural representation in general generative outputs.

\section{Experiment Design for Cultural Bias Evaluation}
\label{sec:3}
\subsection{Experimental Setup}\label{sec:3.1}
We design two complementary studies: a T2I study for bias via image generation, and an I2I study for competence via editing culturally salient details.
We evaluate across six geographically diverse countries (China, India, Kenya, Korea, Nigeria, U.S.), selected to span multiple regions while keeping expert evaluation feasible, addressing underrepresentation in training data~\cite{nayak2024benchmarking}. The overall framework is illustrated in Figure~\ref{fig:framework}.

\begingroup
\captionsetup[figure]{skip=\baselineskip}

\begin{figure}[t]
    \centering
    \includegraphics[width=\linewidth]{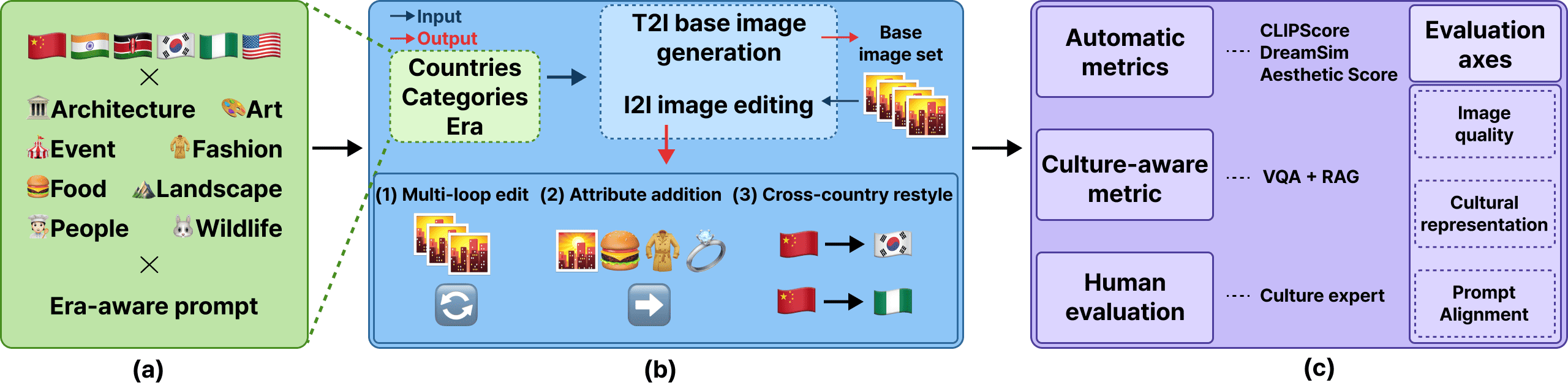}
    \caption{Overall framework overview. (a) Schema inputs: six countries, eight categories, and three era-aware prompts. (b) Experimental pipeline: T2I base generation and three I2I editing studies. (c) Multi-layered evaluation: integrating automatic, culture-aware metrics, and human evaluation.}
    \label{fig:framework}
\end{figure}
\endgroup

\textbf{Category schema.}
Building on prior works~\cite{kannen2024beyond, romero2024cvqa}, our schema defines eight top-level categories (36 subcategories) covering both tangible artifacts (e.g., architecture, cuisine) and intangible practices (e.g., rituals, festivals), roles, and settings. This granular system probes cultural competence by enabling era-aware prompting and controlled cross-country comparisons to evaluate breadth and depth, rather than aesthetics alone systematically. The full schema is shown in Table~\ref{tab:category}.

\textbf{Models.}
We conducted T2I and I2I experiments using state-of-the-art open-source models. Models were selected to capture capability diversity and ensure reproducibility. To our knowledge, none of these models is officially reported by their developers as targeting any specific region. The full list of models is summarized in Table~\ref{tab:models_t2i_i2i} with all settings detailed in Appendix~\ref{app:A}.

\begingroup
\setlength{\textfloatsep}{\baselineskip}
\captionsetup[table]{aboveskip=\baselineskip, belowskip=\baselineskip, skip=\baselineskip}

\begin{table}[!b]
  \centering
  \caption{Models used in T2I and I2I experiments.}
  \label{tab:models_t2i_i2i}
  \setlength{\tabcolsep}{6pt}
  \renewcommand{\arraystretch}{1.15}
  \newcolumntype{Y}{>{\centering\arraybackslash}X} 
  \begin{tabularx}{\linewidth}{@{}YYY@{}}
    \toprule
    \textbf{Family} & \textbf{T2I} & \textbf{I2I} \\
    \midrule
    Stable Diffusion~\cite{esser2024scaling} & Stable Diffusion 3.5 Medium & Stable Diffusion 3.5 Medium \\
    FLUX.1~\cite{labs2025flux} & FLUX.1 [schnell] fp8 & FLUX.1 Kontext [dev] \\
    HiDream~\cite{cai2025hidream} & HiDream-I1-Dev & HiDream-E1.1 \\
    Qwen-Image~\cite{wu2025qwen} & Qwen-Image & Qwen-Image-Edit \\
    NextStep~\cite{team2025nextstep} & NextStep-1-Large & NextStep-1-Large-Edit \\
    \bottomrule
  \end{tabularx}
\end{table}
\endgroup

\begingroup
\setlength{\textfloatsep}{\baselineskip}
\captionsetup[table]{aboveskip=\baselineskip, belowskip=\baselineskip}

\begin{table}[!b]
  \caption{Category schema used in our experiments.}
  \label{tab:category}
  \centering
  \setlength{\tabcolsep}{3.5pt}        
  \renewcommand{\arraystretch}{0.9}    
  \newcolumntype{C}{>{\raggedright\arraybackslash}p{0.18\linewidth}}
  \newcolumntype{Y}{>{\raggedright\arraybackslash}X}

  \begin{tabularx}{\linewidth}{@{} C Y C Y @{}}
    \toprule
    \textbf{category} & \textbf{subcategories} & \textbf{category} & \textbf{subcategories} \\
    \midrule
    \textbf{Architecture} & House, Landmark
                 & \textbf{Landscape} & City, Countryside, Nature \\
    \textbf{Art}          & Dance, Painting, Sculpture
                 & \textbf{Fashion}   & Accessories, Clothing, Makeup \\
    \textbf{Food}         & Beverage, Dessert, Main dish, Snack, Staple food
                 & \textbf{Wildlife}  & Animal, Plant \\
    \addlinespace[2pt]

    \textbf{People}
      & \multicolumn{3}{>{\raggedright\arraybackslash}p{\dimexpr\linewidth-0.18\linewidth\relax}}{%
          Daily life, Athlete, Bride and groom, Celebrity, Chef, Doctor, Farmer,
          Model, President, Soldier, Student, Teacher%
        } \\
    \bottomrule
  \end{tabularx}
\end{table}
\endgroup

\textbf{Evaluation metrics.}
We evaluate models using two complementary dimensions: general-purpose and culture-aware. For general-purpose, we adopt CLIPScore, DreamSim, and Aesthetic Score to measure semantic alignment, cumulative edit distance, and visual quality. For culture-aware evaluation, we extend VQA- and retrieval-augmented generation (RAG) -based approaches~\cite{romero2024cvqa, lewis2020retrieval} by retrieving Wikipedia context via FAISS~\cite{li2025ravenea} and generating yes/no questions and answer with Qwen2.5-0.5B and Qwen2-VL-7B~\cite{qwen2025qwen25technicalreport, team2024qwen2}. Human ratings serve as the primary reference, enabling analysis of consistency and divergence between automatic and culture-aware metrics.

\noindent\textbf{Prompt design.}
To disentangle temporal stylization from cultural identity, we use three prompt modes for each country/category/subcategory:
\begin{itemize}[leftmargin=*,labelindent=0pt,itemsep=0pt,topsep=2pt,parsep=0pt]
    \item Traditional: \texttt{``Traditional \{subcategory\} in \{Country\}, photorealistic.''}
    \item Modern: \texttt{``Modern \{subcategory\} in \{Country\}, photorealistic.''}
    \item Era-agnostic: \texttt{``\{Subcategory\} in \{Country\}, photorealistic.''}
\end{itemize}
This era-aware prompting allows us to examine how cultural depictions vary across specified and unspecified eras, revealing era-sensitive biases.

\subsection{Text-to-Image (T2I) Experiment}
\label{sec:3.2}
We construct a standardized base image set to expose model-internal cultural priors under controlled text-only prompting. For each country and subcategory, we issue era-aware prompts (traditional, modern, and era-agnostic) while fixing all sampling parameters within each model family (Table~\ref{tab:models_t2i_i2i}, T2I column). This yields a comparable collection of generations across all settings, which is subsequently used for distributional analyses, traditional-modern scoring, and seeding the I2I editing experiments in Section~\ref{sec:3.3}. Fixing prompts and sampling settings isolates differences attributable to the models rather than to prompt phrasing or parameter drift.

\subsection{Image-to-Image (I2I) Experiment}
\label{sec:3.3}
We evaluate cultural editing competence with three experimental designs using open-source I2I models--- (Table~\ref{tab:models_t2i_i2i}, I2I column)---reflecting contemporary editing paradigms while remaining reproducible.

\subsubsection{Multi-loop edit: cultural consistency under iterative edits}
\label{sec:3.3.1}
To minimize cross-model domain shifts, base images are drawn from a single T2I model family. We then apply the instruction \texttt{``Change the image to represent \{era\} \{subcategory\} in \{Country\}''} for five successive rounds. This iterative experiment tests stability and path dependence in cultural editing---whether iterations converge toward culturally faithful attributes or drift by amplifying stereotypes.

\subsubsection{Attribute addition: compositional cultural grounding}
\label{sec:3.3.2}
To isolate composition effects, we begin with a neutral canvas: a genderless green mannequin on a white background devoid of cultural cues. For each country/model, a fixed five-step sequence controls (1) background, (2) local-script text rendering, (3) food, (4) clothing, and (5) traditional accessories. This sequence allows us to assess, within a single experiment, the model’s understanding across multiple facets commonly flagged as challenging in cultural benchmarks. Full stepwise prompts are provided in the Appendix~\ref{app:F.1}.

\subsubsection{Cross-country restyle: style transfer across countries}
\label{sec:3.3.3}
We use T2I base images from HiDream-I1-Dev (selected for strong T2I fidelity) and test whether I2I models can restyle an entire scene into a target country’s aesthetic while preserving the subject’s identity and pose. We employ a single minimal prompt---\texttt{``Transform this image into the \{CountryAdj\} style.''}---and report results under this condition; in preliminary trials, adding era/context/attribute cues did not materially improve cultural plausibility, so we do not vary prompt specificity further. This setup supports (i) qualitative audits of cultural appropriateness and context/era consistency and (ii) inspection of edit stability across multi-loop steps.

\subsection{Expert Human Evaluation}
\label{sec:3.4} 
Our human evaluation is intentionally minimal to highlight a core research question: can existing automatic metrics substitute for human evaluation on cultural content? Specifically, we compare CLIPScore against human prompt-alignment ratings and Aesthetic Score against human-rated image quality.
We complement automatic metrics with an expert-group human evaluation, run on our web platform (ECB Human Survey) under a unified protocol. Raters evaluate only their own country (emic expertise), ensuring cultural authenticity in judgments. For each prompt, four candidates (step base/1/3/5) are displayed side-by-side. Raters assign two scores: 1-5 Likert scores for \emph{Image Quality} and \emph{Cultural Representation}. These two components are averaged to form the Human Quality Score (HQS). Raters also select \emph{Best/Worst}, following recent practices in cultural assessment and editing evaluation~\cite{nayak2025culturalframes,li2025balancing}. Image sets span eight categories across five models (Table~\ref{tab:models_t2i_i2i}). Each rater completes six tasks: five I2I-loop evaluations and one attribute addition evaluation. Expert raters require insider knowledge and language proficiency. Operational details, including the distribution of our expert raters, are provided in Appendix~\ref{app:B}.

\section{Results}
\label{sec:4}

\begingroup
\captionsetup[figure]{skip=\baselineskip}

\begin{figure}[t]
    \centering
    \includegraphics[width=\linewidth]{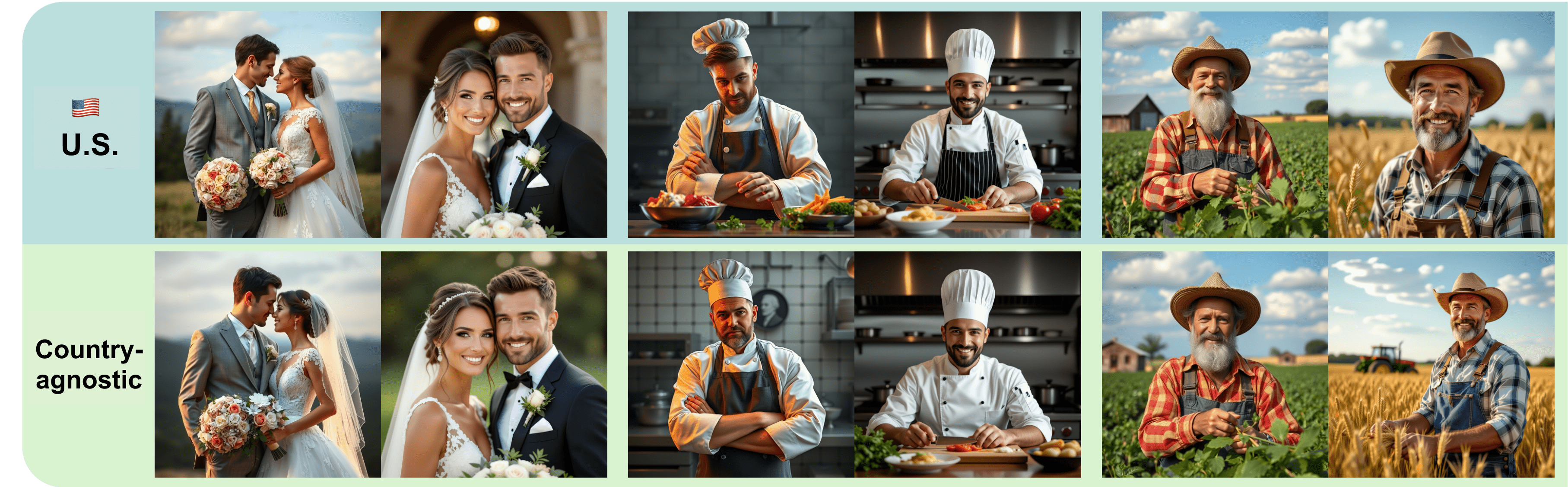}
    \caption{Comparative samples for U.S. (top) and country-agnostic (bottom) prompts across two models. Left image: FLUX.1 [schnell] fp8; right image: HiDream-I1-Dev. Within each model, panels show (from left) \textit{Bride and groom}, \textit{Chef}, and \textit{Farmer}. The close correspondence between rows illustrates the US-like default of country-agnostic prompts.}
    \label{fig:lean}
\end{figure}
\endgroup

\subsection{Text-to-Image (T2I)}
\label{sec:4.1}

We use a unified embedding backbone across analyses. Let \(x_i \in \mathbb{R}^d\) denote the CLIP image embedding of sample \(i\) generated by one of \(M{=}5\) T2I models, and let \(c(i)\in\mathcal{C}\) be its country tag (\emph{country-agnostic} means no explicit tag). When clustering is required, we perform model-wise PCA to two components and run \(k\)-means with \(K_m\) clusters. All mathematical definitions (cluster-proportion vectors, Jensen-Shannon divergence (JSD), the proximity metrics, and the traditional-modern leaning score with its summaries) are given in Appendix~\ref{app:C}; we reference equations there as needed.

\subsubsection{Distributional Proximity and Cultural Defaults}
\label{sec:4.1.1}

For each model \(m\) and country \(c\), we compute cluster proportions via Equation~\ref{eq:pvec} and measure pairwise country similarity using Equation~\ref{eq:h} with \(\mathrm{JSD}\) from Equation~\ref{eq:jsd}; model-averaged proximity is then given by Equation~\ref{eq:hbar}.
Across all five models, the strongest proximity is U.S.\ \(\leftrightarrow\) country-agnostic (\(\bar h{=}0.892\), 95\% CI \([0.844,0.931]\)), followed by China \(\leftrightarrow\) Korea (\(0.888\), \([0.835,0.932]\)) and Kenya \(\leftrightarrow\) Nigeria (\(0.864\), \([0.811,0.911]\)). A meta-analysis indicates negligible between-model heterogeneity (\(\tau^2 \approx 0\) except Kenya--Nigeria \(\tau^2{\approx}8.5\times10^{-5}\)), indicating stable effects. These findings support a robust U.S.-like default for country-agnostic prompts and regional clustering in East Asia and Sub-Saharan Africa. Model-level nearest-neighbor votes confirm China--Korea and Kenya--Nigeria as mutual neighbors, while India often aligns with U.S./country-agnostic, suggesting partial assimilation. See Figure~\ref{fig:lean} for a visual comparison: U.S. and country-agnostic prompts produce almost identical samples, qualitatively presenting the U.S.-like default.

\subsubsection{Traditional--Modern Leaning under Country-agnostic Prompts}
\label{sec:4.1.2}

Image-level traditional-modern scores (Equation~\ref{eq:si}) are computed per model, aggregated to country means via Equation~\ref{eq:sc}, and their significance is evaluated using within-category permutation with BH-FDR across countries. 
The cross-country dispersion of \(\{\bar s_c\}\) is significant in every model (four models: \(p{=}0.001\); Qwen-Image: \(p{=}0.002\)), indicating systematic differences under country-agnostic prompts. The dominant pattern is a consistent \emph{modern} lean for the U.S. and country-agnostic across all five models (\(\bar s_c{<}0\), per-model \(p{\le}0.006\), FDR-significant). Other effects are small and model-dependent: Kenya is traditional-leaning only in FLUX.1 [schnell] fp8 (\(p{=}0.001\), FDR-significant), while India, Nigeria, Korea, and China are mixed or non-significant (typically \(q{>}0.1\)).
Category composition largely explains the country means: \textit{clothing} and \textit{sport} consistently yield modern (negative) scores across models and countries---especially for the U.S. and country-agnostic---whereas \textit{house}, \textit{religious ritual}, and \textit{landmark} generally yield traditional (positive) scores for most countries but often flip to modern for the U.S. and country-agnostic. These patterns indicate that the modern default arises from systematic topic/style composition rather than sampling noise. Per-country \(p\)-values and BH-FDR-adjusted \(q\)-values appear in Appendix~\ref{app:C}, Table~\ref{tab:t2i_country_sig_all}.

\subsection{Image-to-Image (I2I)}
\label{sec:4.2}

\subsubsection{Multi-Loop Edit}\label{sec:4.2.1}

\begingroup
\captionsetup[figure]{skip=\baselineskip}
\begin{figure}[t]
    \centering
    \includegraphics[width=1.0\linewidth]{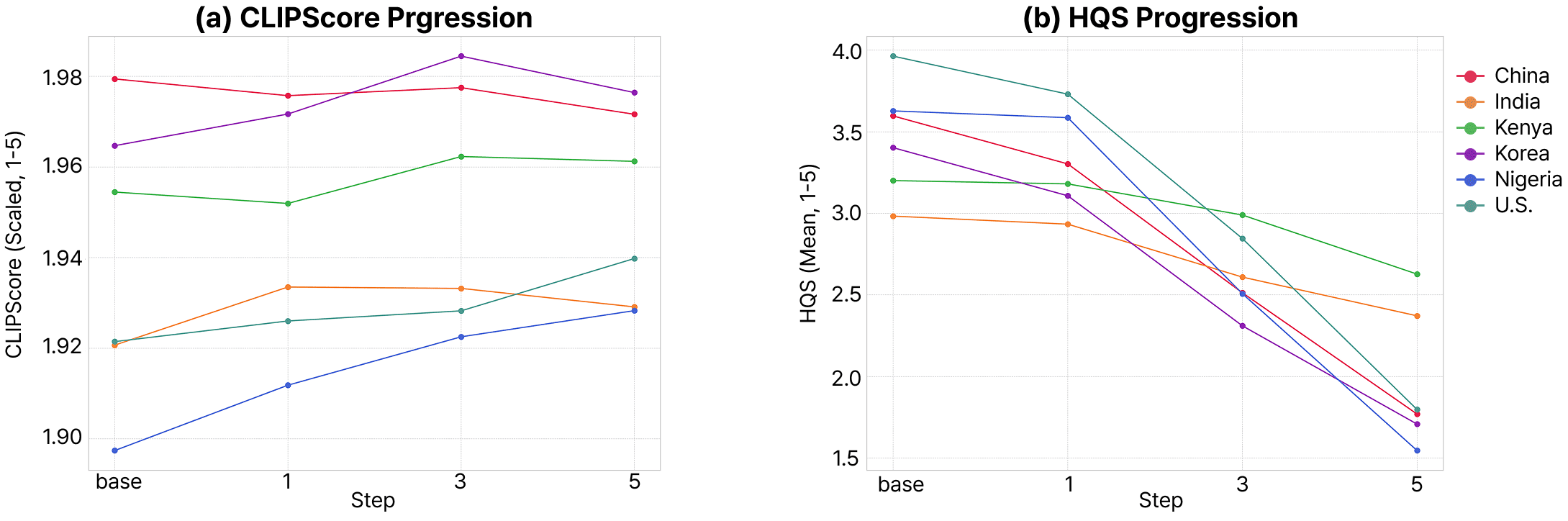}
    \caption{Divergence between Automated and Human Judgment in Iterative Editing. (a) CLIPScore trajectories remain largely stable or modestly increase from the base step to step 5. (b) Human Quality Score (HQS) sharply declines across all countries. This pronounced divergence highlights the failure of traditional automatic metrics to track the perceptible cultural degradation that human raters consistently penalize.}
    \label{fig:multi-loop-edit}
\end{figure}
\endgroup


\begingroup
\captionsetup[figure]{skip=\baselineskip}
\begin{figure}[t]
    \centering
    \includegraphics[width=1.0\linewidth]{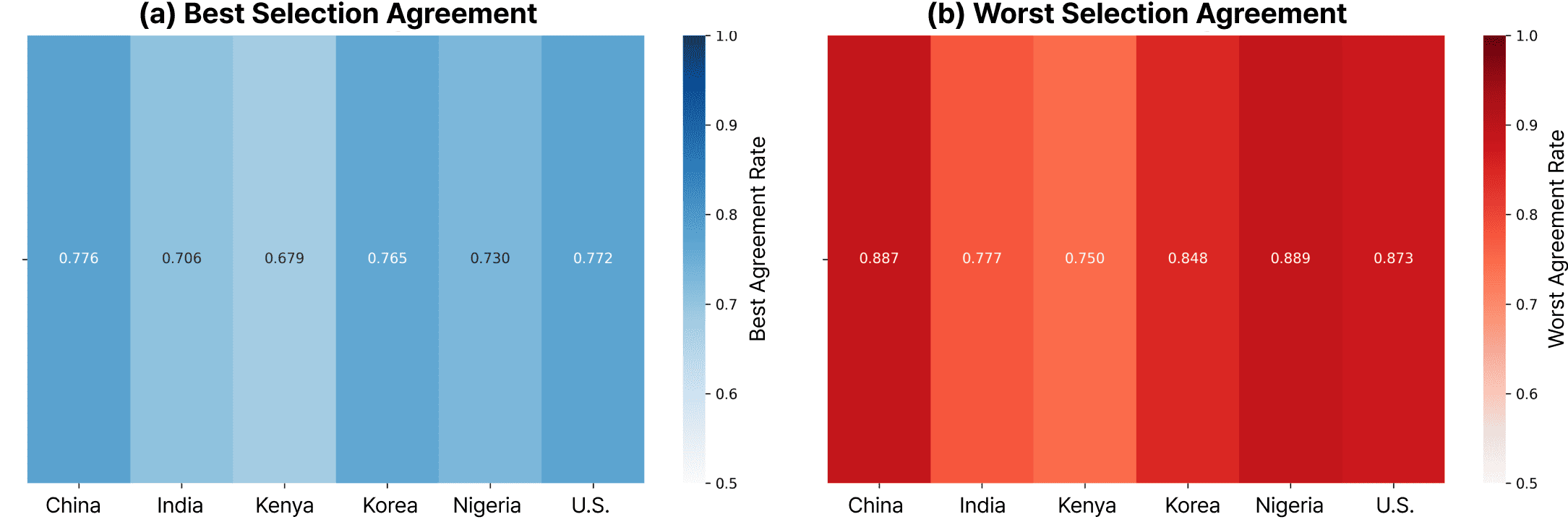}
    \caption{Alignment of the Culture-aware Metric with Human Judgment (All Models Averaged). (a) The agreement rate for \textit{Best Selection} is high across all countries, averaging 73.8\%. (b) The agreement rate for \textit{Worst Selection} is consistently higher, averaging 83.7\%. This high alignment demonstrates that our extended culture-aware metric successfully tracks human preference for unedited states and penalizes pronounced cultural erosion. Detailed stepwise score changes are provided in the Appendix~\ref{app:D.1}.}
    \label{fig:best-worst}
\end{figure}
\endgroup
Following the multi-loop experiment in Section~\ref{sec:3.3.1}, we report country-level averages across models for five successive I2I edits. Our findings reveal a substantial gap between automatic metrics and human judgment. As shown in Figure~\ref{fig:multi-loop-edit}, CLIPScore remains largely flat or slightly increases across steps, whereas human-related HQS sharply declines for most model-country pairs. This divergence indicates that with continued edits, culturally salient cues---such as context and era---gradually erode, yielding images that appear more aligned with prompts yet remain culturally distorted.

Other automatic metrics, including Aesthetic Score and DreamSim, exhibit modest or decreasing trends but fail to capture culture-specific degradation. In contrast, our culture-aware metric shows a stepwise decline consistent with HQS, demonstrating its sensitivity to cultural loss. Figure~\ref{fig:best-worst} further illustrates this alignment: the agreement with human selection reaches 73.8\% for best images and 83.7\% for worst images.

Overall, while traditional automatic metrics remain stable or even improve despite perceptible cultural deterioration, our culture-aware metric aligns more closely with human perception. These results suggest that culturally sensitive editing becomes unreliable under iterative I2I, where standard metrics may obscure degradation that both human evaluators and our extended metric reveal. Detailed per-model and per-country analyses are provided in Appendix~\ref{app:D}, and additional qualitative examples are included in Appendix~\ref{app:F.2}.

\subsubsection{Attribute Addition}
\label{sec:4.2.2}
Following the attribute-addition protocol in Section~\ref{sec:3.3.2}, we test whether models can compose culture-specific elements step by step. We sequentially add five attributes---background, local-script text, food, clothing, and accessories---and visualize representative progressions in Figure~\ref{fig:attribute}. For evaluation, the expert group in Section~\ref{sec:3.4} rated each step on three dimensions (Likert): image quality, prompt alignment, and cultural fidelity. The results are as follows:
The sequential attribute-addition task quantified compositional failure, revealing a consistent decline in image quality (IQ) due to the compounding of edits. Crucially, while FLUX.1 Kontext [dev] demonstrated superior overall IQ, Qwen-Image-Edit achieved higher prompt alignment in complex compositional tasks (Food/Clothing), suggesting a fundamental tradeoff between strict object control and visual quality. This analysis highlights two critical failure points: the Text attribute (sharp alignment declines from gibberish) and the final Accessory attribute (lowest IQ scores), indicating persistent difficulty rendering complex, culturally specific details.

\begingroup
\setlength{\floatsep}{2pt plus 1pt minus 1pt} 
\setlength{\textfloatsep}{\baselineskip}
\captionsetup[figure]{aboveskip=\baselineskip}

\begin{figure}[t]
    \centering
    \includegraphics[width=1.0\linewidth]{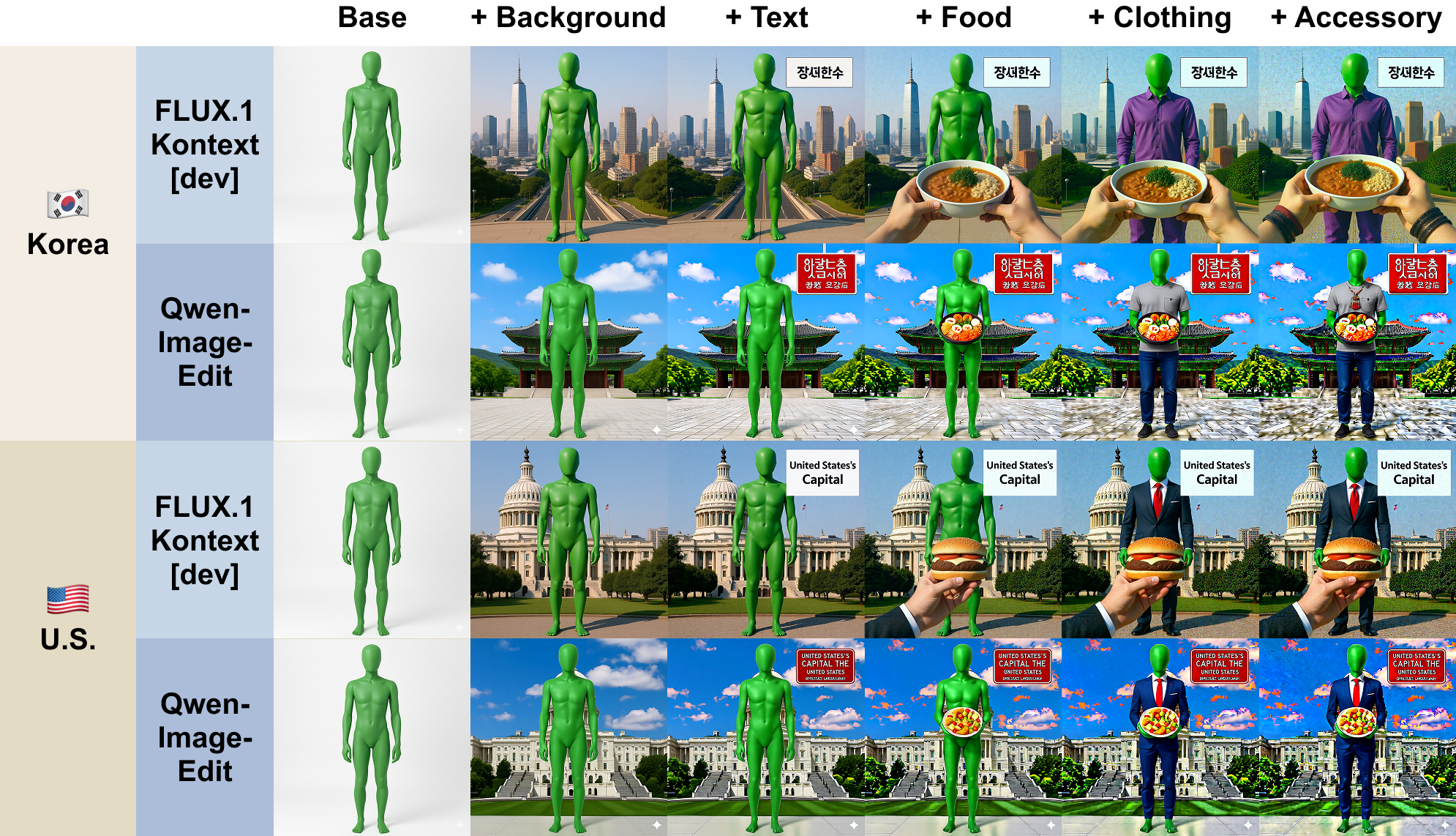}
    \caption{Representative stepwise attribute additions for Korea and the U.S. Columns progress from the base image to background, text, food, clothing, and accessory, each added cumulatively. Rows denote the model; the top block uses Korea prompts and the bottom block U.S. prompts.}
    \label{fig:attribute}
\end{figure}

\begin{figure}[t]
    \centering
    \includegraphics[width=1.0\linewidth]{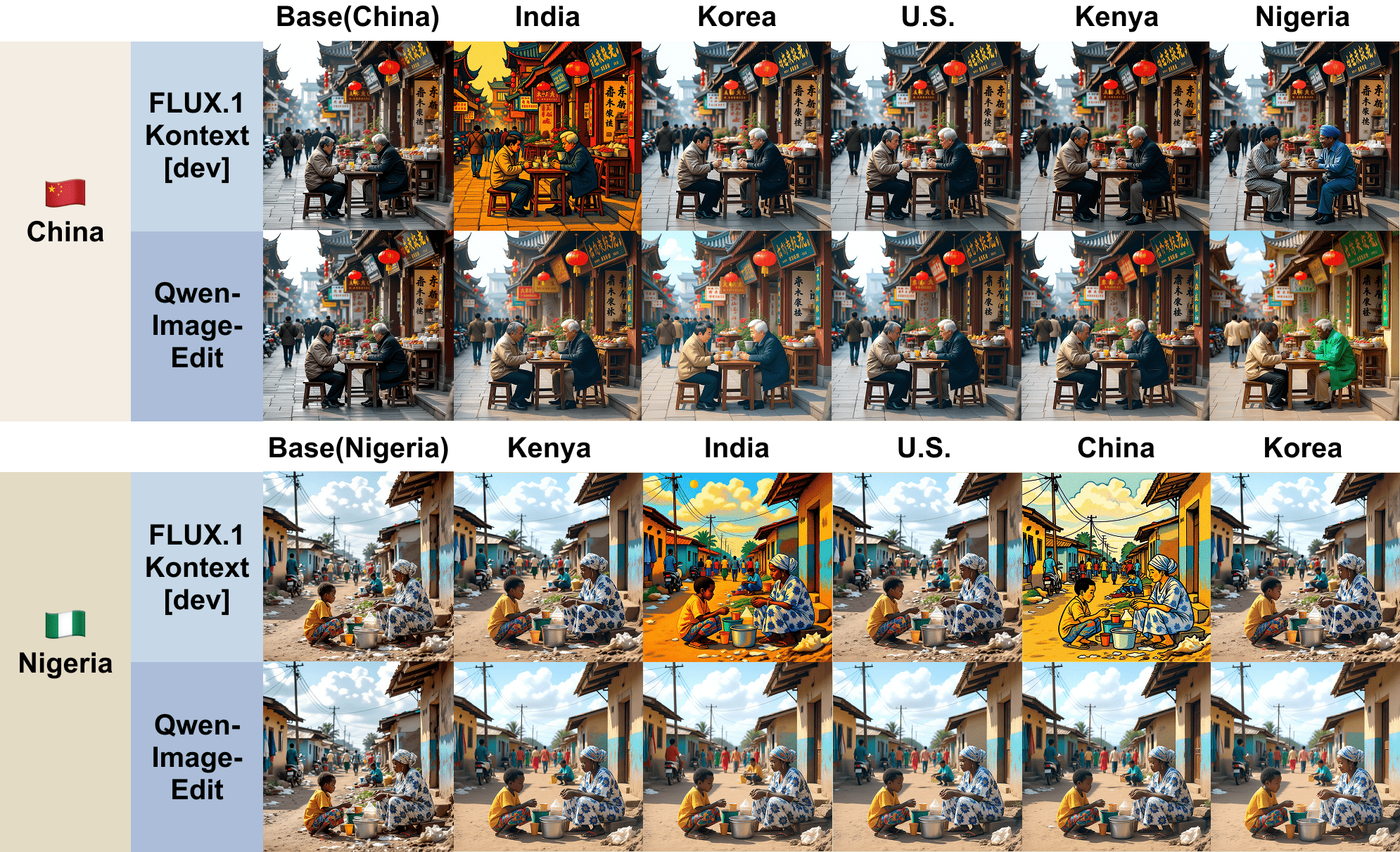}
    \caption{Representative cross-country restyling results for China and Nigeria; columns are ordered by geographic proximity to the base country---closest on the left, farthest on the right. Rows indicate the model; the top block uses a China base image and the bottom block a Nigeria base image.}
    \label{fig:cross-country}
\end{figure}
\endgroup

\subsubsection{Cross-Country Restylization}
\label{sec:4.2.3}
Using the single minimal prompt (Section~\ref{sec:3.3.3}), we qualitatively assess two I2I models; Figure~\ref{fig:cross-country} shows the outcomes. Both models preserve layout, composition, and pose across multi-loop steps, but culture-relevant details drift: localized attributes (attire, objects, vernacular architecture, signage) remain under-edited or degrade. Qwen-Image-Edit often substitutes symbolic or palette shifts associated with the target country for true localized changes, yielding surface “signals” rather than context- or era-consistent edits. When targeting non-U.S. countries (particularly those from Africa and Asia), subjects frequently retain their original phenotype (race/skin tone), indicating weak identity adaptation. We also observe a style asymmetry: non-U.S. targets are commonly rendered in drawing/painting styles, whereas U.S. targets remain photorealistic. Overall, under minimal prompting, current editing models favor superficial markers and structural conservation over culture- and era-faithful transformations. Additional qualitative examples are included in Appendix~\ref{app:F.2}.

\section{Discussion and Limitations}
\label{sec:disc-limit}
\textbf{Discussion.}
Our experiments show substantial cultural bias in current image generators, with three dominant patterns. (i) U.S.-like default: country-agnostic prompts produce U.S.-like, modern-leaning outputs (Figure~\ref{fig:lean}); stable regional pairings (e.g., China-Korea; Kenya-Nigeria) indicate that category mix and style priors---not noise---drive cross-country gaps. (ii) Metric-human gap in iterative I2I: conventional metrics (e.g., CLIPScore) stay flat or rise while human-rated cultural quality drops (Figure~\ref{fig:multi-loop-edit}); our culture-sensitive metric tracks human Best/Worst choices (Figure~\ref{fig:best-worst}), revealing metric drift. (iii) Shortcut editing: models ``signal'' culture via superficial cues (flag overlays, palette swaps), preserve source identity when targeting non-U.S. countries, and render non-U.S. scenes in painterly styles while keeping U.S. scenes photorealistic (Figure~\ref{fig:attribute}, \ref{fig:cross-country}). These cultural effects co-occur with occupation-level demographic skews---male dominance in several roles, female skew in caregiving/aesthetic roles, and light-tone prevalence under neutral prompts (see Appendix~\ref{app:G}).

\textbf{Limitations.}
Our I2I experiments start from within-family T2I baselines to isolate intra-model behavior and reduce cross-family confounds, which limits claims about transfer across model families. We use generic prompts (no per-model engineering), so results reflect out-of-the-box behavior rather than tuned best case. Scale (five models, six countries) is bounded by human-evaluation cost and compute constraints---multi-loop editing and large-batch inference require substantial GPU resources. A further limitation is the unit of analysis: we use country-level labels as a proxy for “culture” (sovereign states, e.g., U.S., China), not subnational units. Prior work cautions that aligning culture with geopolitical borders obscures within-country heterogeneity, minority communities, and transnational/diaspora groups~\cite{nayak2025culturalframes, bayramli2025diffusion}; this is salient for the U.S. (immigration-shaped, strong regional variation) and for China (many officially recognized minority groups). We also acknowledge the risk of cultural essentialism; to mitigate this, we use era-aware prompts, attribute-addition tests, and flexible contextual cues rather than prescriptive symbols. Nevertheless, intra-country diversity, the potential for stereotyping, and concerns about cultural offensiveness warrant further study. Finally, automated components used in evaluation (e.g., VQA/LLM-assisted steps) can introduce their own biases, so results should be interpreted with care~\cite{bayramli2025diffusion}. Future work should adopt finer-grained and potentially multi-label groupings (subnational regions, language/community groups) and audit automated tools, as recommended by prior studies~\cite{nayak2025culturalframes, bayramli2025diffusion}.

\section{Conclusion and Future Work}
\label{sec:conclusion}
We presented a structured evaluation of cultural bias across T2I and I2I using six countries, an 8-category/36-subcategory schema, era-aware prompts (\textit{traditional}/\textit{modern}/\textit{era-agnostic}), and three protocols (Multi-Loop Edit, Attribute Addition, Cross-Country Restylization). Our open platform pairs standard automatic metrics with a culture-sensitive metric and a web-based human study workflow, enabling triangulation and model-level diagnostics. Empirically, country-agnostic prompting defaults to U.S.-like, modern-leaning outputs; iterative I2I can erode cultural fidelity while conventional metrics obscure the decline; and editing pipelines often rely on superficial cues rather than culture-consistent, context-preserving changes.
Progress hinges on two practical directions. \emph{Finer granularity:} report beyond country tags---include subnational (state/province/city) and community/language groups---and use simple stratified reporting so diverse communities are not collapsed into a single label. \emph{Stronger training/data signals:} curate balanced datasets; add era-aware, context-preserving objectives to both generation and editing; and explicitly penalize shortcut cues that fake culture (e.g., flag overlays, generic palettes, style flips that ignore people/objects/setting). We release all images, prompts, and configurations to support reproducible follow-up studies along these directions.

\begin{ack}
We would like to express our sincere gratitude to the following contributors for their valuable support and collaboration throughout this work: Parth Maheshwari, Abdullahi Abdulrahman, Ge Fang, Dusillah Dullo, Alice Etori, Abdullahi Adavize Ismaila, SoHee Yoon, Jamin Lee, Ikbum Park, Taeseo Kim, and Hyowon Choi.
This work is in part supported by NSF IIS-2112633 and J.P. Morgan.
Minki Hong, Sieun Choi, and Jihie Kim are supported by the MSIT (Ministry of Science and ICT) of Korea under the Global Research Support Program in the Digital Field (RS-2024-00426860) and the Artificial Intelligence Convergence Innovation Human Resources Development (IITP-2026-RS-2023-00254592), supervised by the IITP (Institute for Information \& Communications Technology Planning \& Evaluation).
\end{ack}

\newpage

{\small
\bibliographystyle{unsrt}
\bibliography{references}
}

\clearpage
\appendix
\section*{Appendix}

\section{Model Configuration}
All experiments were conducted on a workstation equipped with an AMD Threadripper Pro 5955WX processor (16 cores, 4.0 GHz) and 128 GB of RAM. The system includes two NVIDIA RTX 4090 GPUs and a 2 TB NVMe SSD for the operating system and storage. The operating system is Ubuntu 22.04, managed with the Lambda Stack for CUDA, cuDNN, TensorFlow, and PyTorch.
\label{app:A}
\subsection{T2I \& I2I Model Configuration}\label{app:A.1}

\begin{table}[h]
\centering
\caption{Model parameters and average per-image runtime (bs=1, 1024×1024): T2I generators (left) and I2I editors (right).}
\vspace{\baselineskip} 
\label{tab:model-config}

\begingroup
\setlength{\tabcolsep}{3pt} 
\footnotesize               

\begin{subtable}[t]{0.48\linewidth}
\centering
\renewcommand{\arraystretch}{1.15}
\begin{tabularx}{\linewidth}{@{}>{\raggedright\arraybackslash}X c c@{}}
\toprule
\textbf{Model} & \textbf{Params} & \textbf{Avg.\ time} \\
\midrule
Stable Diffusion 3.5 Medium & 2.5B & 8s      \\
FLUX.1 [schnell] fp8        & 12B  & 2s      \\
HiDream-I1-Dev              & 17B  & 25s     \\
Qwen-Image                  & 20B  & 2m 5s   \\
NextStep-1-Large            & 15B  & 1m 30s  \\
\bottomrule
\end{tabularx}
\end{subtable}
\hfill
\begin{subtable}[t]{0.48\linewidth}
\centering
\renewcommand{\arraystretch}{1.15}
\begin{tabularx}{\linewidth}{@{}>{\raggedright\arraybackslash}X c c@{}}
\toprule
\textbf{Model} & \textbf{Params} & \textbf{Avg.\ time} \\
\midrule
Stable Diffusion 3.5 Medium & 2.5B & 5s     \\
FLUX.1 Kontext [dev]        & 12B  & 39s    \\
HiDream-E1.1                & 8B   & 1m 11s \\
Qwen-Image-Edit             & 20B  & 2m 5s  \\
NextStep-1-Large-Edit       & 15B  & 2m 49s \\
\bottomrule
\end{tabularx}
\end{subtable}

\endgroup
\end{table}

\subsection{Model Configuration using Culture-Aware Metric}\label{app:A.2}
For culture-aware metric, following prior VQA- and RAG-based approaches~\cite{romero2024cvqa, lewis2020retrieval}, we extend these implementations by retrieving Wikipedia context for each cultural category using FAISS index~\cite{li2025ravenea}, prompting Qwen2.5-0.5B-Instruct~\cite{qwen2025qwen25technicalreport} to generate contextual yes/no questions (including negative checks), and using Qwen2-VL-7B-Instruct~\cite{team2024qwen2} to provide image and context answers. We derive two axes---image quality and cultural representation---and conduct group comparisons to select best/worst images with concise rationales. While QA metrics such as Precision, Recall, and F1 are reported for auditing purposes, human ratings remain our primary reference.

\section{Human Evaluation Platform and Protocol}
\label{app:B}
This appendix expands Section~\ref{sec:3.4} with expert-only operational details of the \textsc{ECB Human Survey} web platform, task flow, rater recruitment, and quality control. Our study uses a unified protocol: for each prompt, four candidates (base, step~1, step~3, step~5) are displayed side-by-side in randomized order; raters assign three 1--5 Likert scores---\emph{Image Quality}, \emph{Prompt Alignment}, and \emph{Cultural Representation}---and select \emph{Best/Worst} with an optional one-line rationale.

\subsection{Platform Overview and UI Snapshots}\label{app:B.1}
Figure~\ref{fig:appA-survey-ui} shows representative screens from the survey platform: consent/IRB gating, the participant dashboard, the multi-loop edit interface, and the attribute-addition interface used for stepwise cultural edits.

\begingroup
\captionsetup[figure]{skip=\baselineskip}

\begin{figure}[ht]
    \centering
    \begin{subfigure}[t]{0.48\linewidth}
        \includegraphics[width=\linewidth]{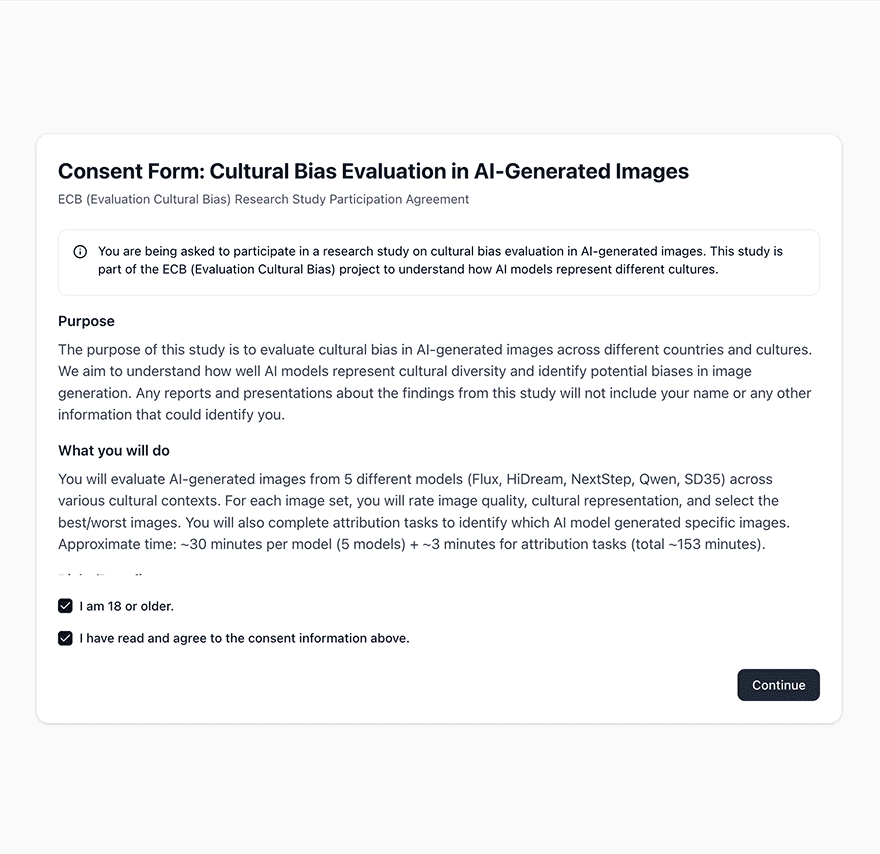}
        \caption{Consent/IRB gating. Purpose, eligibility, and informed consent.}
        \label{fig:appA-irb}
    \end{subfigure}\hfill
    \begin{subfigure}[t]{0.48\linewidth}
        \includegraphics[width=\linewidth]{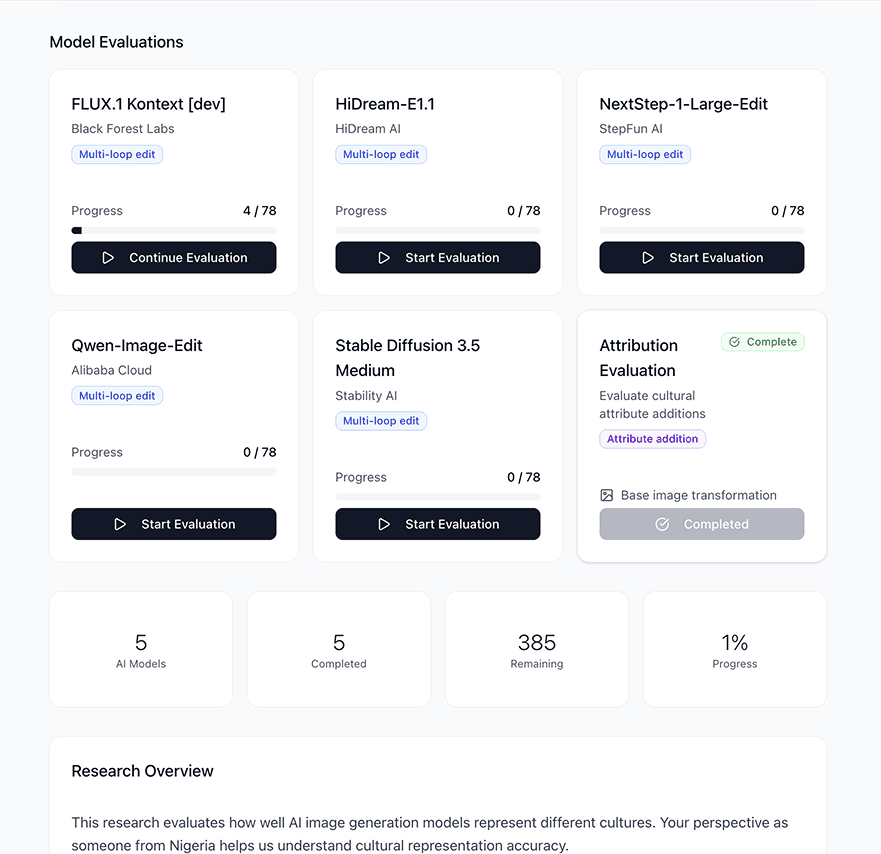}
        \caption{Participant dashboard. Country selection, model cards, progress, and survey guideline.}
        \label{fig:appA-dashboard}
    \end{subfigure}

    \vspace{0.75em}
    \begin{subfigure}[t]{0.48\linewidth}
        \includegraphics[width=\linewidth]{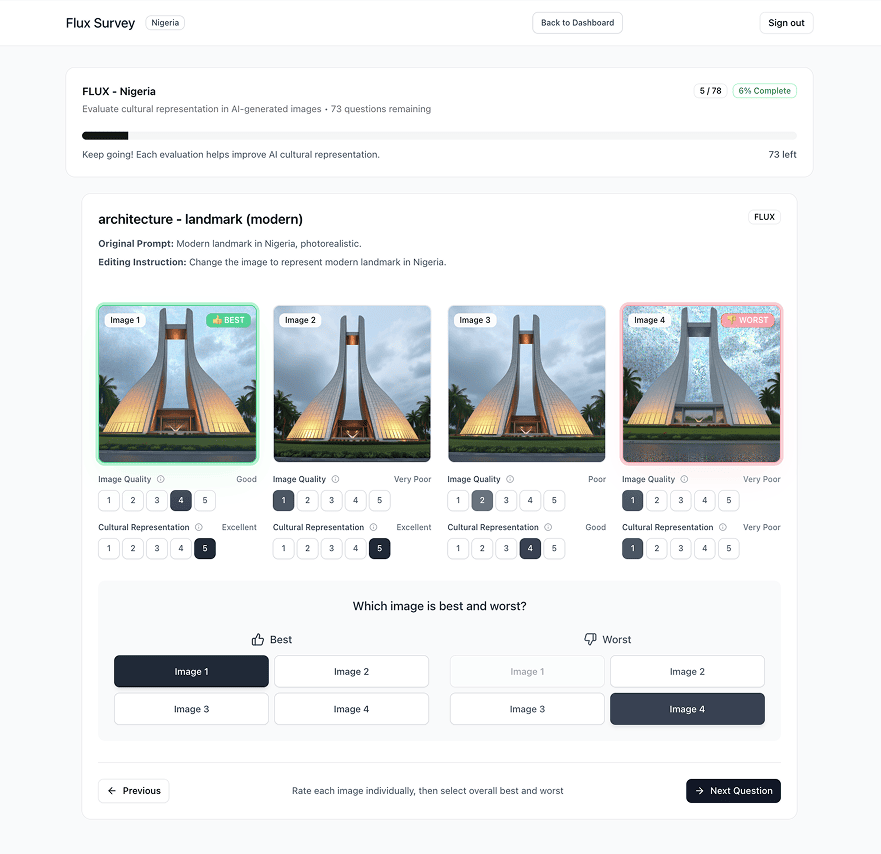}
        \caption{A survey participation screen featuring an image, a Likert scale, and a best/worst selection task.}
        \label{fig:appA-multiloop}
    \end{subfigure}\hfill
    \begin{subfigure}[t]{0.48\linewidth}
        \includegraphics[width=\linewidth]{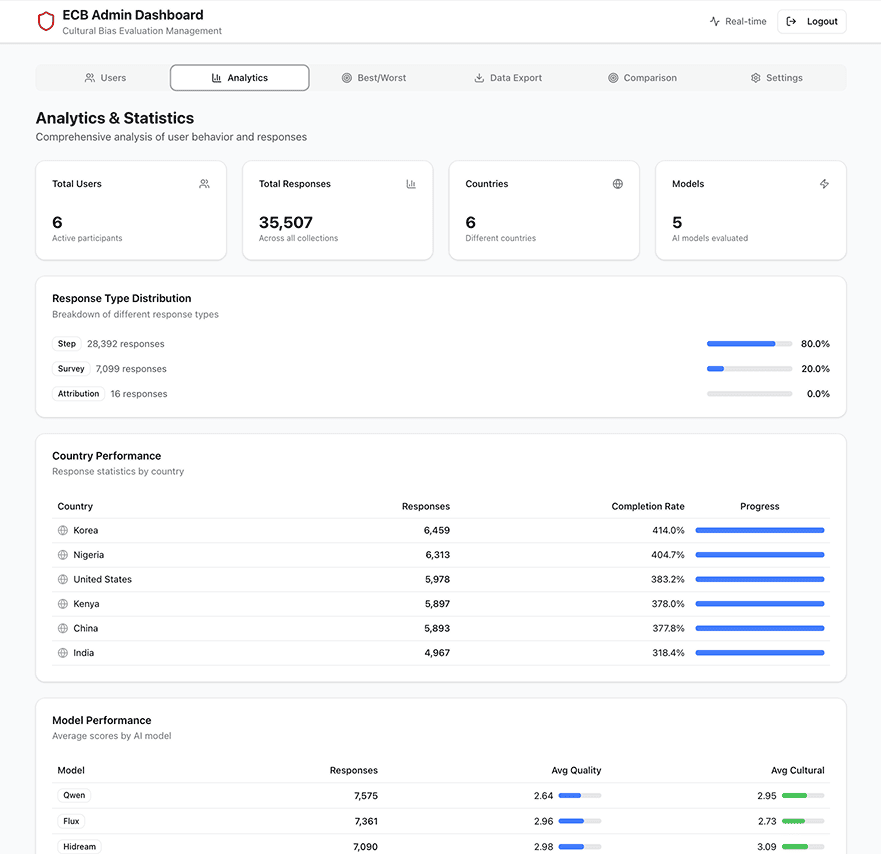}
        \caption{An admin dashboard for viewing the overall progress of survey participants and their statistics.}
        \label{fig:appA-attr}
    \end{subfigure}

    \caption{ECB Human Survey UI snapshots.
    (a) consent/IRB gating; (b) participant dashboard; (c) survey participation screen; (d) admin dashboard}
    \label{fig:appA-survey-ui}
\end{figure}
\endgroup

\subsection{Task Flow and Randomization}\label{app:B.2}
Sessions begin with consent gating and a short prescreen. Tasks are drawn from a country-specific pool and randomized at three levels to reduce presentation bias: (i) prompt order, (ii) image position within each row, and (iii) model order across tasks. Raters complete five multi-loop edit evaluations (one per model) and one attribute-addition evaluation, using the same 1--5 scales and Best/Worst selection.

\subsection{Expert Recruitment and Emic Expertise}\label{app:B.3}
We conducted an \textbf{expert-only} study. A total of \textbf{17 domain experts} participated---\textbf{3 per country} across China, Kenya, Korea, Nigeria, and the United States, and \textbf{2 from India}. Experts evaluated \emph{only their own country} (\emph{emic} expertise), leveraging insider cultural understanding (context, symbols, style, era, regional specificity). Eligibility required insider knowledge of the target country (residency or sustained lived experience) and proficiency in relevant language(s). All participants confirmed no conflicts of interest with model development or data curation.

\subsection{Quality Control and Ethics}\label{app:B.4}
We employ consent gating, randomized presentation, and embedded gold items (country-specific binary checks and sanity items). We flag submissions with inconsistent answers, string-identical rationales, or excessive speed, and review flags before exclusion. Participation ensures anonymity, fair compensation, and the right to withdraw. The platform logs anonymized ratings, Best/Worst choices, rationales, and per-row hashes for auditability.

\subsection{Reproducibility Artifacts}\label{app:B.5}
We release UI templates, randomization seeds, and scripts to reproduce the survey and aggregation pipeline. Static UI snapshots used in Figure~\ref{fig:appA-survey-ui} are included in the repository under \texttt{figures/appendix\_survey/}.

\section{Supplementary details for T2I analysis}\label{app:C}

\begin{equation}\label{eq:pvec}
\mathbf{p}^{(m)}_{c}=(p^{(m)}_{c,0},\dots,p^{(m)}_{c,K_m-1}),\quad
p^{(m)}_{c,k}=\frac{\#\{i: c(i)=c,\,k(i)=k\}}{\#\{i: c(i)=c\}},\quad
\sum_{k}p^{(m)}_{c,k}=1.
\end{equation}\vspace{-12pt}

\begin{equation}\label{eq:jsd}
\mathrm{JSD}(p,q)=\tfrac12\mathrm{KL}(p\|m)+\tfrac12\mathrm{KL}(q\|m),\quad m=\tfrac12(p{+}q).
\end{equation}\vspace{-12pt}

\begin{equation}\label{eq:h}
h^{(m)}_{ab}=\frac{2\,\cos(\mathbf{p}^{(m)}_{a},\mathbf{p}^{(m)}_{b})\,[1-\mathrm{JSD}(\mathbf{p}^{(m)}_{a},\mathbf{p}^{(m)}_{b})]}
{\cos(\mathbf{p}^{(m)}_{a},\mathbf{p}^{(m)}_{b})+[1-\mathrm{JSD}(\mathbf{p}^{(m)}_{a},\mathbf{p}^{(m)}_{b})]}.
\end{equation}\vspace{-12pt}

\begin{equation}\label{eq:hbar}
\bar h_{ab}=\tfrac{1}{M}\sum_{m=1}^{M} h^{(m)}_{ab}.
\end{equation}\vspace{-12pt}

\begin{equation}\label{eq:si}
s_i=\cos(x_i,\mu_{\text{trad}}(g(i)))-\cos(x_i,\mu_{\text{mod}}(g(i))).
\end{equation}\vspace{-12pt}

\begin{equation}\label{eq:sc}
\bar s_c=\tfrac{1}{n_c}\sum_{i\in\mathcal{I}_c}s_i,\qquad
\mathrm{SE}(\bar s_c)=\tfrac{\mathrm{sd}(\{s_i:i\in\mathcal{I}_c\})}{\sqrt{n_c}}.
\end{equation}\vspace{+12pt}

\begingroup
\scriptsize
\setlength{\tabcolsep}{3pt}
\setlength{\LTcapwidth}{\linewidth}
\setlength{\LTleft}{0pt}
\setlength{\LTright}{0pt}

\setlength{\LTpre}{\baselineskip}
\setlength{\LTpost}{\baselineskip}

\begin{longtable}{@{}%
  >{\raggedright\arraybackslash}p{0.27\linewidth}
  >{\raggedright\arraybackslash}p{0.17\linewidth}
  S[table-format=+1.2]
  S[table-format=1.2]
  S[table-format=1.2]
  S[table-format=1.2]
  S[table-format=1.2]
  S[table-format=1.2]
  >{\raggedright\arraybackslash}p{0.12\linewidth}
  @{}}
\caption{Traditional-modern leaning by country across T2I models.
Columns: \emph{Mean margin} ($\bar s_c$; $<$0 = modern, $>$0 = traditional), \emph{SE} (standard error across images),
\emph{cos(trad)} / \emph{cos(mod)} = cosine similarity to the traditional/modern anchors,
$p$ = permutation $p$-value, $q_{\mathrm{FDR}}$ = BH-FDR within each model across countries,
and \emph{Lean} = sign-based label. All values rounded to two decimals.}%
\label{tab:t2i_country_sig_all}\\
\noalign{\vskip\normalbaselineskip}
\toprule
Model & Country & {Mean margin} & {SE} & {cos(trad)} & {cos(mod)} & {$p$} & {$q_{\mathrm{FDR}}$} & Lean \\
\midrule
\endfirsthead
\toprule
Model & Country & {Mean margin} & {SE} & {cos(trad)} & {cos(mod)} & {$p$} & {$q_{\mathrm{FDR}}$} & Lean \\
\midrule
\endhead
\midrule
\multicolumn{9}{r}{Continued on next page}\\
\midrule
\endfoot
\bottomrule
\endlastfoot

\multirow{7}{=}{Stable Diffusion 3.5 Medium}
 & China            & +0.01 & 0.02 & 0.82 & 0.81 & 0.67 & 0.79 & traditional \\
 & India            & +0.03 & 0.02 & 0.84 & 0.81 & 0.18 & 0.42 & traditional \\
 & Kenya            & +0.01 & 0.01 & 0.81 & 0.80 & 0.82 & 0.82 & traditional \\
 & Korea            & -0.03 & 0.01 & 0.80 & 0.83 & 0.25 & 0.44 & modern \\
 & Nigeria          & -0.02 & 0.01 & 0.83 & 0.84 & 0.55 & 0.77 & modern \\
 & United States    & -0.06 & 0.02 & 0.73 & 0.79 & 0.00 & 0.00 & modern \\
 & Country-agnostic & -0.07 & 0.02 & 0.78 & 0.85 & 0.00 & 0.00 & modern \\
\midrule
\multirow{7}{=}{FLUX.1 [schnell] fp8}
 & China            &  0.00 & 0.02 & 0.84 & 0.84 & 0.91 & 0.91 & traditional \\
 & India            & +0.02 & 0.01 & 0.85 & 0.82 & 0.10 & 0.17 & traditional \\
 & Kenya            & +0.04 & 0.01 & 0.81 & 0.77 & 0.00 & 0.00 & traditional \\
 & Korea            & -0.02 & 0.01 & 0.83 & 0.85 & 0.25 & 0.35 & modern \\
 & Nigeria          & +0.01 & 0.02 & 0.83 & 0.82 & 0.47 & 0.54 & traditional \\
 & United States    & -0.05 & 0.01 & 0.75 & 0.80 & 0.00 & 0.00 & modern \\
 & Country-agnostic & -0.04 & 0.01 & 0.81 & 0.85 & 0.01 & 0.01 & modern \\
\midrule
\multirow{7}{=}{HiDream-I1-Dev}
 & China            &  0.00 & 0.02 & 0.81 & 0.80 & 0.99 & 0.99 & traditional \\
 & India            & +0.01 & 0.02 & 0.82 & 0.81 & 0.89 & 0.99 & traditional \\
 & Kenya            & +0.02 & 0.02 & 0.77 & 0.76 & 0.67 & 0.99 & traditional \\
 & Korea            & -0.01 & 0.02 & 0.80 & 0.81 & 0.78 & 0.99 & modern \\
 & Nigeria          & -0.02 & 0.02 & 0.82 & 0.84 & 0.64 & 0.99 & modern \\
 & United States    & -0.10 & 0.02 & 0.73 & 0.83 & 0.00 & 0.01 & modern \\
 & Country-agnostic & -0.08 & 0.02 & 0.75 & 0.83 & 0.01 & 0.02 & modern \\
\midrule
\multirow{7}{=}{Qwen-Image}
 & China            & -0.01 & 0.02 & 0.84 & 0.84 & 0.74 & 0.75 & modern \\
 & India            & +0.01 & 0.02 & 0.84 & 0.82 & 0.56 & 0.75 & traditional \\
 & Kenya            & +0.01 & 0.02 & 0.82 & 0.81 & 0.54 & 0.75 & traditional \\
 & Korea            & -0.02 & 0.02 & 0.81 & 0.83 & 0.34 & 0.75 & modern \\
 & Nigeria          & -0.01 & 0.02 & 0.79 & 0.80 & 0.75 & 0.75 & modern \\
 & United States    & -0.05 & 0.02 & 0.75 & 0.80 & 0.01 & 0.02 & modern \\
 & Country-agnostic & -0.06 & 0.02 & 0.79 & 0.84 & 0.00 & 0.02 & modern \\
\midrule
\multirow{7}{=}{NextStep-1-Large}
 & China            & +0.02 & 0.01 & 0.87 & 0.85 & 0.27 & 0.47 & traditional \\
 & India            & +0.03 & 0.02 & 0.84 & 0.82 & 0.13 & 0.30 & traditional \\
 & Kenya            &  0.00 & 0.01 & 0.81 & 0.82 & 0.83 & 0.95 & modern \\
 & Korea            & +0.01 & 0.02 & 0.85 & 0.85 & 0.78 & 0.95 & traditional \\
 & Nigeria          &  0.00 & 0.02 & 0.84 & 0.83 & 0.95 & 0.95 & traditional \\
 & United States    & -0.07 & 0.02 & 0.78 & 0.85 & 0.00 & 0.00 & modern \\
 & Country-agnostic & -0.06 & 0.02 & 0.77 & 0.84 & 0.00 & 0.00 & modern \\
\end{longtable}
\endgroup

\section{Detailed Model-Country Analysis}\label{app:D}
In this section, we utilize the model family names found in Table~\ref{tab:models_t2i_i2i}.

\subsection{Best/Worst Selection Patterns}\label{app:D.1}
Figure~\ref{fig:best-worst-patterns} visualizes best/worst selection patterns across all model-country pairs, contrasting human selections with those from our culture-aware metric used in Section~\ref{sec:4.2.1}.

\begingroup
\captionsetup[figure]{skip=\baselineskip}
\begin{figure}[ht]
    \centering
    \includegraphics[width=1.0\linewidth]{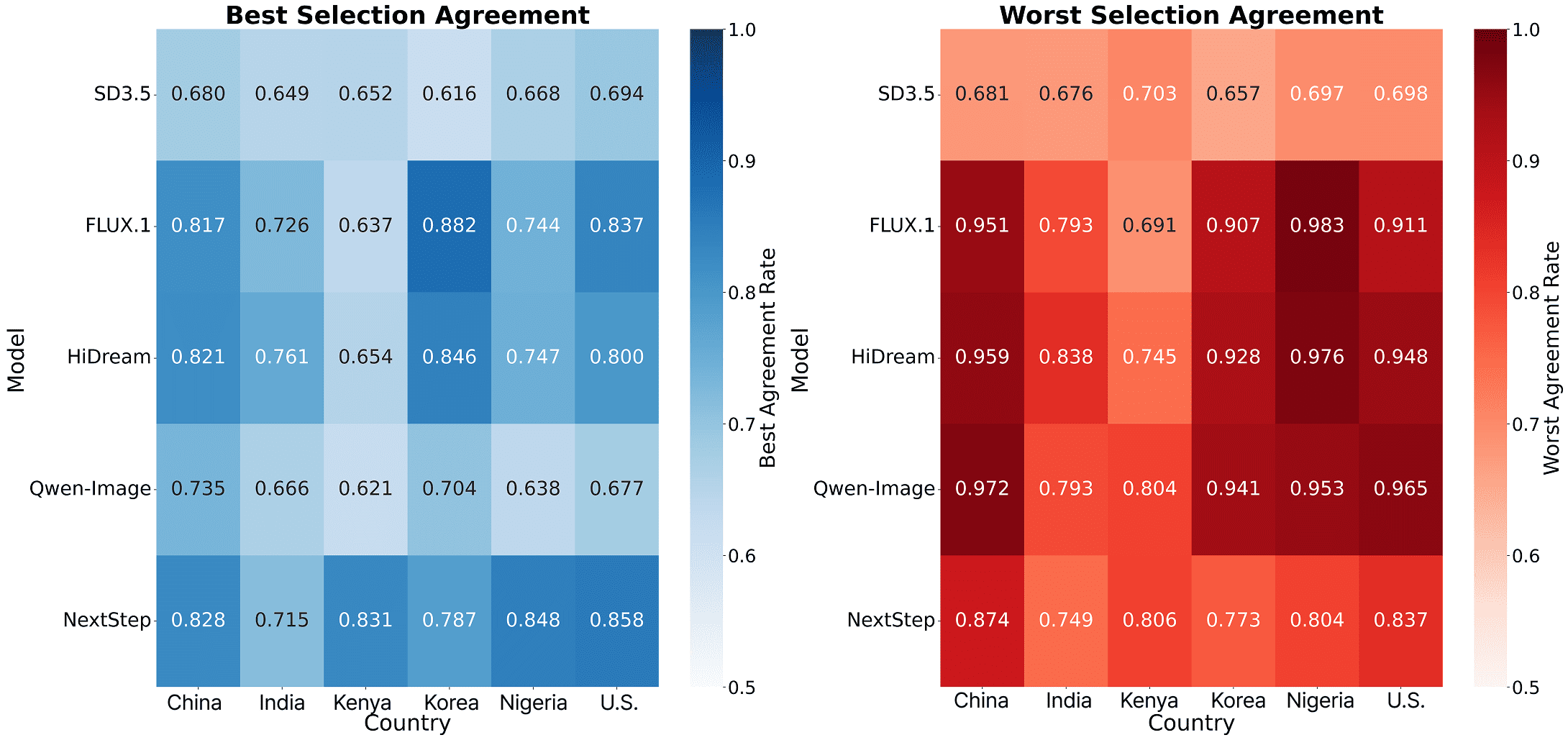}
    \caption{Alignment of the human selection with our culture-aware metric by model and country.}
    \label{fig:best-worst-patterns}
\end{figure}
\endgroup

\subsection{Step-Wise Quality Degradation Analysis}\label{app:D.2}

\subsubsection{CLIPScore Changes by Model-Country}
\begingroup
\captionsetup[figure]{skip=\baselineskip}
\begin{figure}[ht]
    \centering
    \includegraphics[width=0.95\linewidth]{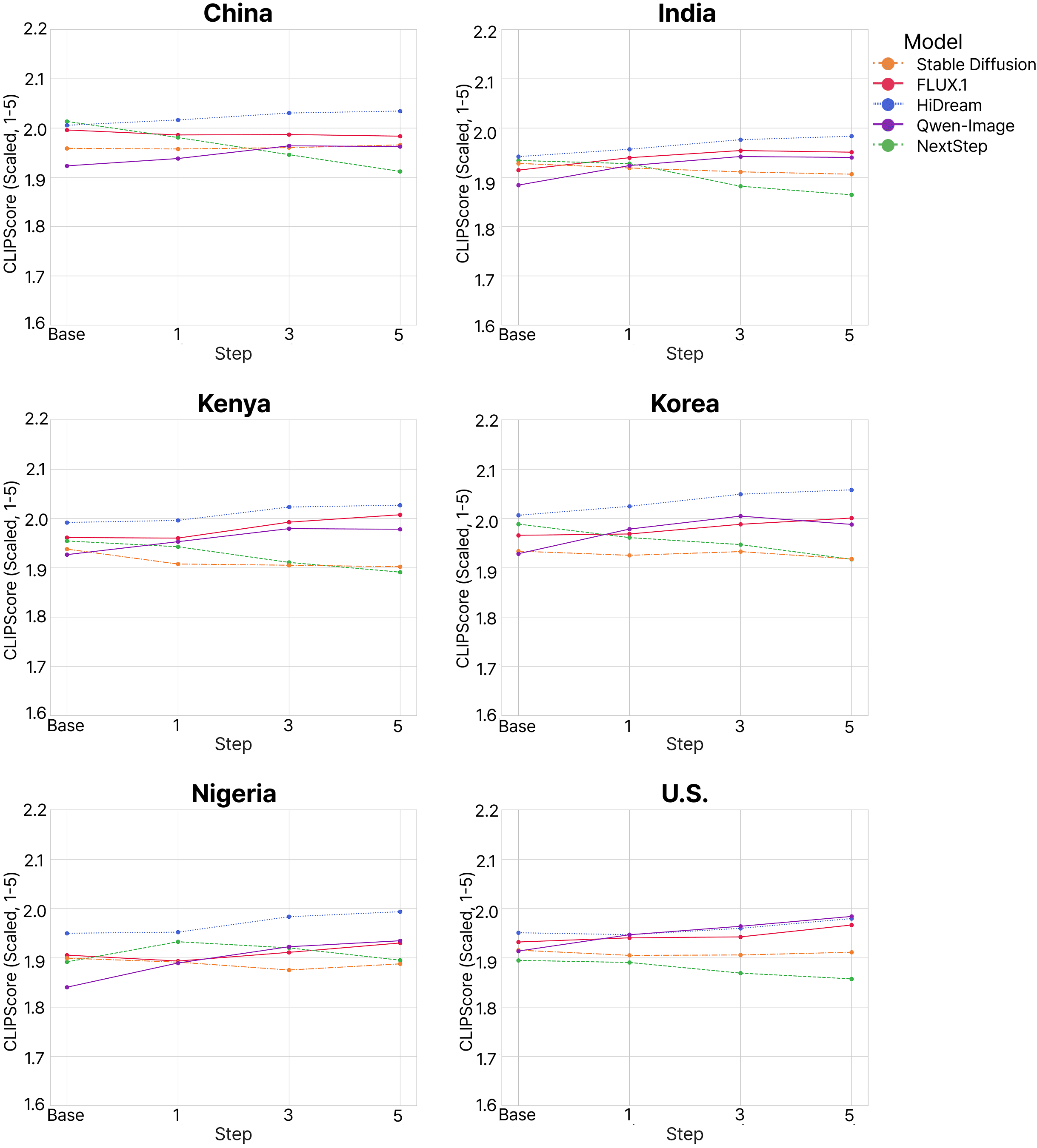}
    \caption{Average CLIPScore Progression Across Iterative I2I Steps by Country and Model.}
    \label{fig:app-clip}
\end{figure}
\endgroup
Figure~\ref{fig:app-clip} shows the change in CLIPScore by country. Across all six countries, the CLIPScore exhibits a generally flat or slightly increasing trend over the five I2I steps. This stability or modest ascent supports our main result's claim that traditional automatic metrics fail to register the cultural degradation observed by human evaluators. The CLIPScore trajectories for most models---Stable Diffusion (SD3.5), HiDream, Qwen-Image, and NextStep---typically cluster within a narrow range (1.90 to 2.05). The FLUX.1 model often registers the highest CLIPScore across steps in most countries, such as Korea and the United States (U.S.), suggesting a superior ability to maintain prompt-image alignment during iterative editing. However, this does not correlate with human-perceived cultural quality. While the overall trend is flat, minor country-specific nuances are observable: In Korea and the U.S., nearly all models demonstrate a consistent, albeit minor, increase in CLIPScore, culminating in the highest average scores among the six countries. In Nigeria, the CLIPScore for models like HiDream and NextStep shows virtually no change (a flat line), highlighting the metric's insensitivity to any potential cultural drift in this context. In China and Kenya, the CLIPScore exhibits slight volatility, with minor drops or gains between steps (e.g., NextStep in China), but overall remains within a small range, confirming the metric's stability.

Table~\ref{tab:appD-clip} reports the change in CLIPScore from base to step 5, showing a mean increase of 0.7\% across all pairs, with a range of -5.1\% to +5.1\%.

\begin{table}[ht]
\centering
\caption{CLIPScore changes by model-country (base image to step~5). `Change' is the change in CLIPScore, and `Final' is the final CLIPScore.}

\resizebox{\textwidth}{!}{%
    \begin{tabular}{l *{5}{c c}}
    \toprule
    \multirow{2}{*}{\textbf{Country}} & \multicolumn{2}{c}{\textbf{SD3.5}} & \multicolumn{2}{c}{\textbf{FLUX.1}} & \multicolumn{2}{c}{\textbf{HiDream}} & \multicolumn{2}{c}{\textbf{Qwen-Image}} & \multicolumn{2}{c}{\textbf{NextStep}} \\
    \cmidrule(lr){2-3} \cmidrule(lr){4-5} \cmidrule(lr){6-7} \cmidrule(lr){8-9} \cmidrule(lr){10-11}
    & Change (\%) & Final & Change (\%) & Final & Change (\%) & Final & Change (\%) & Final & Change (\%) & Final \\
    \midrule
    China   & +0.4 & 1.97 & -0.6 & 1.98 & +1.4 & 2.03 & +2.0 & 1.96 & -5.0 & 1.91 \\
    India   & -1.1 & 1.91 & +1.9 & 1.95 & +2.1 & 1.98 & +3.0 & 1.94 & -3.6 & 1.86 \\
    Kenya   & -1.8 & 1.90 & +2.4 & 2.01 & +1.8 & 2.03 & +2.7 & 1.98 & -3.2 & 1.89 \\
    Korea   & -0.8 & 1.92 & +1.8 & 2.00 & +2.6 & 2.06 & +3.1 & 1.99 & -3.6 & 1.92 \\
    Nigeria & -0.6 & 1.89 & +1.3 & 1.93 & +2.2 & 1.99 & +5.1 & 1.93 & +0.2 & 1.90 \\
    U.S.    & -0.2 & 1.91 & +1.8 & 1.97 & +1.5 & 1.98 & +3.7 & 1.98 & -2.0 & 1.86 \\
    \bottomrule
    \end{tabular}%
}
\label{tab:appD-clip}
\end{table}

\subsubsection{Human Quality Score (HQS) Changes by Model-Country}

\begingroup
\captionsetup[figure]{skip=\baselineskip}
\begin{figure}[ht]
    \centering
    \includegraphics[width=0.95\linewidth]{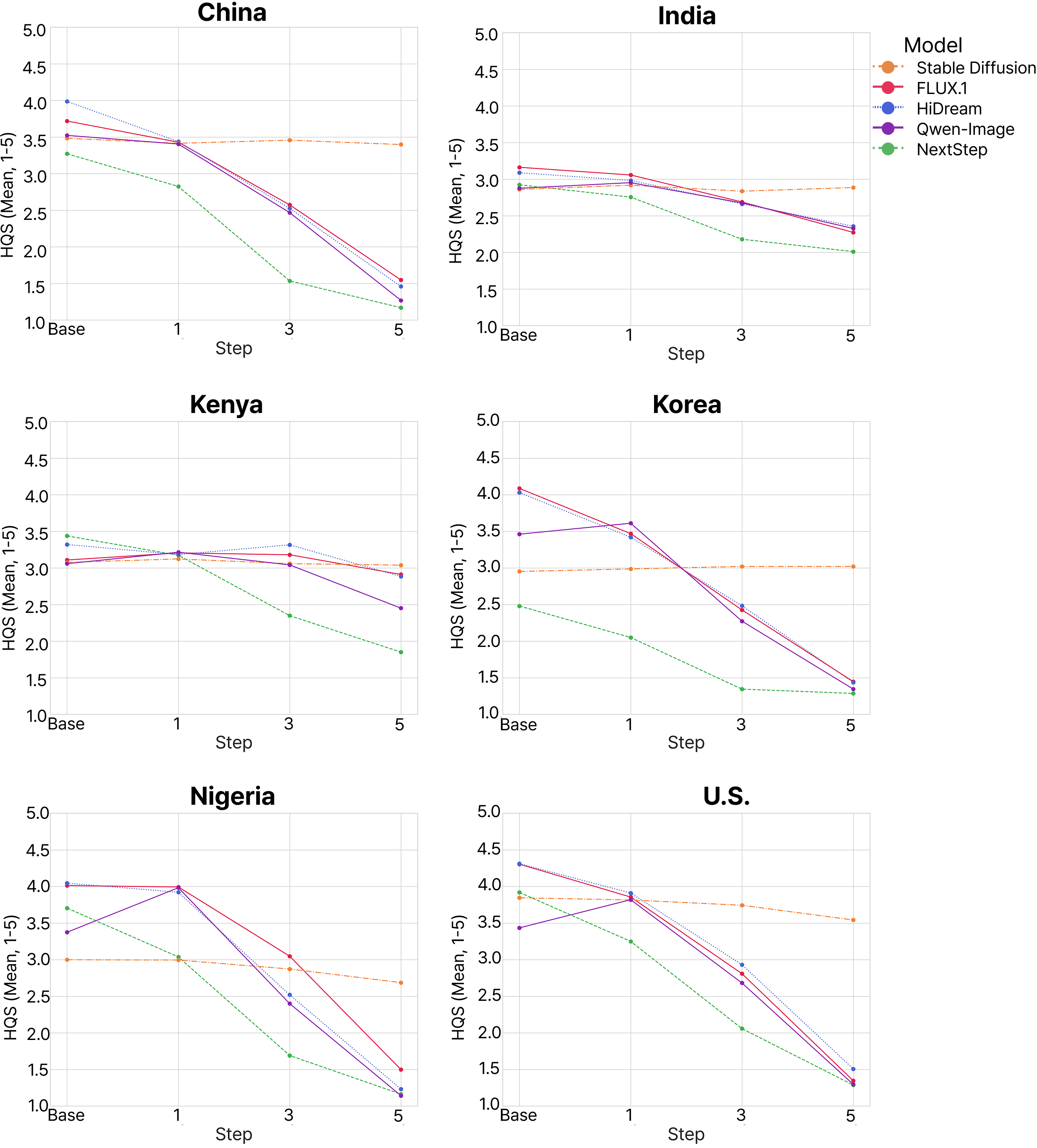}
    \caption{Average HQS Progression Across Iterative I2I Steps by Country and Model.}
    \label{fig:app-human}
\end{figure}
\endgroup

Figure~\ref{fig:app-human} shows the change in HQS by country. In contrast to the flat or increasing trend of the CLIPScore, the Average HQS demonstrates a significant and steep decline across all six countries for the majority of models, particularly after step 1. This deterioration strongly aligns with human perception that iterative editing leads to substantial cultural and aesthetic degradation. For most high-performing models---Stable Diffusion, HiDream, Qwen-Image, and NextStep---the HQS drops from an initial score near 4.0 (or higher) down to a final score between 1.0 and 2.5 by step 5. The most pronounced decline is observed in Korea and the U.S., where models like Stable Diffusion and HiDream drop to scores near 1.0, indicating near-total failure in cultural or aesthetic quality by the final edit. FLUX.1, while still showing a measurable decline, consistently maintains the highest HQS at step 5 across all countries. This result suggests that although cultural erosion is present, FLUX.1 is more robust at preserving key image elements over iterative edits compared to its counterparts. The steepest drop for most models often occurs between step 1 and step 3, signaling that the second and third I2I loops introduce critical loss of culturally salient context or coherence.

The consistent and significant decline in HQS across diverse cultural contexts provides compelling evidence that iterative I2I editing is unreliable for preserving human-perceived quality and cultural integrity. Automatic metrics, such as CLIPScore, fail to capture fundamental cultural degradation. Table~\ref{tab:appD-hqs} lists HQS deltas from base to step~5.

\begingroup
\setlength{\textfloatsep}{\baselineskip}
\captionsetup[table]{aboveskip=\baselineskip, belowskip=\baselineskip}

\begin{table}[ht]
\centering
\caption{Human Quality Score changes by model-country (base image to step~5). 'Change' is the change in HQS, and 'Final' is the final HQS.}
\small 
\resizebox{\textwidth}{!}{%
\begin{tabular}{l *{5}{c c}}
\toprule
\multirow{2}{*}{\textbf{Country}} & \multicolumn{2}{c}{\textbf{SD3.5}} & \multicolumn{2}{c}{\textbf{FLUX.1}} & \multicolumn{2}{c}{\textbf{HiDream}} & \multicolumn{2}{c}{\textbf{Qwen-Image}} & \multicolumn{2}{c}{\textbf{NextStep}} \\
\cmidrule(lr){2-3} \cmidrule(lr){4-5} \cmidrule(lr){6-7} \cmidrule(lr){8-9} \cmidrule(lr){10-11}
& Change (\%) & Final & Change (\%) & Final & Change (\%) & Final & Change (\%) & Final & Change (\%) & Final \\
\midrule
China   & -2.4 & 3.40 & -58.4 & 1.55 & -63.4 & 1.46 & -64.0 & 1.27 & -64.2 & 1.17 \\
India   & +0.9 & 2.89 & -28.1 & 2.27 & -23.6 & 2.36 & -19.1 & 2.33 & -31.2 & 2.01 \\
Kenya   & -1.2 & 3.04 & -6.4  & 2.91 & -13.1 & 2.88 & -19.9 & 2.45 & -46.2 & 1.85 \\
Korea   & +2.3 & 3.02 & -64.6 & 1.45 & -64.3 & 1.44 & -61.1 & 1.35 & -48.1 & 1.29 \\
Nigeria & -10.4& 2.69 & -62.6 & 1.50 & -69.5 & 1.23 & -66.1 & 1.14 & -68.6 & 1.16 \\
U.S.    & -7.9 & 3.54 & -68.7 & 1.35 & -65.0 & 1.51 & -62.2 & 1.30 & -67.1 & 1.29 \\
\bottomrule
\end{tabular}
}
\label{tab:appD-hqs}
\end{table}
\endgroup

\subsection{Cultural Bias Analysis}\label{app:D.3}
\subsubsection{Country-Wise Patterns}

Nigeria recorded a severe HQS decline with an average of -55.4\%, suggesting a high vulnerability to degradation under iterative editing. The United States showed a similarly severe HQS decline, averaging -53.9\%, also indicating high susceptibility to quality loss from repetitive edits. India (average -49.5\%), Korea (average -48.2\%), and China (average -46.5\%) all experienced a significant HQS decline, highlighting notable degradation patterns in these countries. In contrast, Kenya recorded a moderate HQS decline averaging -15.8\%, suggesting the quality deterioration was less severe in this context, though some variation across different models was still observed.

\subsubsection{Model-Wise Patterns}

SD3.5 proved to be the most stable, showing a minimal average decline of only -3.0\%; in some cases, like SD3.5-China, it even registered an improvement of +3.2\%. In contrast, all other models exhibited significant degradation. FLUX.1 saw a major drop, averaging -51.4\%. Qwen-Image also experienced a major decline, averaging -52.9\%, closely followed by HiDream with an average degradation of -54.2\%. Finally, NextStep recorded the highest degradation among all models, with a severe average drop of -57.9\%.

\subsection{Agreement Rate Analysis}\label{app:D.4}
\subsubsection{Best Selection Agreement (Base)}

Table~\ref{tab:appB-best} reports the best selection agreement between human judgment and our culture-aware metric for the base image. This metric indicates the models' initial ability to generate culturally preferred images before any iterative editing. Regarding initial performance, FLUX.1 demonstrates the highest initial agreement rates, notably reaching 77.1\% in Korea and 67.4\% in the U.S., suggesting that FLUX.1 exhibits superior fidelity to culturally salient features in the base generation step. HiDream also shows strong initial performance, particularly in Korea (69.7\%). Furthermore, NextStep also registers high agreement, achieving 69.9\% in the U.S. and 68.1\% in China. In contrast, SD3.5 and Qwen-Image generally show the lowest agreement, frequently falling below 45\%. Overall agreement rates show significant national variation, indicating that a model's base-level preference is highly sensitive to the cultural context of the generated image.

\begingroup
\setlength{\textfloatsep}{\baselineskip}
\captionsetup[table]{aboveskip=\baselineskip, belowskip=\baselineskip}

\begin{table}[ht]
\centering
\caption{Best selection agreement at base step.}
\small
\resizebox{\textwidth}{!}{%
\begin{tabular}{l *{5}{c c}}
\toprule
\multirow{2}{*}{\textbf{Country}} & \multicolumn{2}{c}{\textbf{SD3.5}} & \multicolumn{2}{c}{\textbf{FLUX.1}} & \multicolumn{2}{c}{\textbf{HiDream}} & \multicolumn{2}{c}{\textbf{Qwen-Image}} & \multicolumn{2}{c}{\textbf{NextStep}} \\
\cmidrule(lr){2-3} \cmidrule(lr){4-5} \cmidrule(lr){6-7} \cmidrule(lr){8-9} \cmidrule(lr){10-11}
& Agree (\%) & Count & Agree (\%) & Count & Agree (\%) & Count & Agree (\%) & Count & Agree (\%) & Count \\
\midrule
China   & 39.0 & 78  & 62.1 & 152 & 64.3 & 154 & 47.7 & 113 & 68.1 & 160 \\
India   & 34.6 & 47  & 44.7 & 87  & 52.7 & 122 & 34.5 & 79  & 45.3 & 82  \\
Kenya   & 33.2 & 65  & 27.3 & 65  & 30.8 & 72  & 24.5 & 57  & 63.0 & 175 \\
Korea   & 24.6 & 53  & 77.1 & 202 & 69.7 & 169 & 41.4 & 130 & 51.1 & 172 \\
Nigeria & 36.3 & 77  & 48.7 & 151 & 49.6 & 118 & 27.8 & 66  & 64.6 & 179 \\
U.S.    & 39.3 & 89  & 67.4 & 159 & 60.3 & 140 & 36.9 & 87  & 69.9 & 179 \\
\bottomrule
\end{tabular}
}
\label{tab:appB-best}
\end{table}
\endgroup

\subsubsection{Worst Selection Agreement (Step~5)}

Table~\ref{tab:appB-worst} reports the worst selection agreement between human judgment and our culture-aware metric for the step 5 images. The agreement rates for the worst selection at step 5 are significantly higher than the best agreement at the base level, suggesting our metrics are far better at identifying failure cases. FLUX.1, HiDream, and Qwen-Image generally achieve the highest agreement rates, often exceeding 85\%. Notably, HiDream-Nigeria and FLUX.1-Nigeria show near-perfect agreement, indicating that the metrics reliably identified severe degradation in these model-country pairs. Furthermore, agreement rates in India are strong across these models, reaching 90.6\% for FLUX.1 and 67.9\% for HiDream, reinforcing the metric's efficacy in capturing failure modes even in diverse contexts like India. In contrast, SD3.5 exhibits the lowest overall agreement, failing to surpass 45\% in any country, suggesting the degradation pattern of SD3.5 is the least predictable by our metric. The overall high agreement indicates that, by the final stage of editing, the cultural failure modes become so extreme that they are easily detectable by our culture-aware metric, regardless of the specific country context.

\begingroup
\setlength{\textfloatsep}{\baselineskip}
\captionsetup[table]{aboveskip=\baselineskip, belowskip=\baselineskip}

\begin{table}[ht]
\centering
\caption{Worst selection agreement at step~5.}
\small
\resizebox{\textwidth}{!}{%
\begin{tabular}{l *{5}{c c}}
\toprule
\multirow{2}{*}{\textbf{Country}} & \multicolumn{2}{c}{\textbf{SD3.5}} & \multicolumn{2}{c}{\textbf{FLUX.1}} & \multicolumn{2}{c}{\textbf{HiDream}} & \multicolumn{2}{c}{\textbf{Qwen-Image}} & \multicolumn{2}{c}{\textbf{NextStep}} \\
\cmidrule(lr){2-3} \cmidrule(lr){4-5} \cmidrule(lr){6-7} \cmidrule(lr){8-9} \cmidrule(lr){10-11}
& Agree (\%) & Count & Agree (\%) & Count & Agree (\%) & Count & Agree (\%) & Count & Agree (\%) & Count \\
\midrule
China   & 37.3 & 82  & 90.6 & 217 & 91.9 & 216 & 92.4 & 228 & 74.5 & 199 \\
India   & 40.9 & 52  & 57.4 & 116 & 67.9 & 162 & 55.7 & 144 & 49.4 & 98  \\
Kenya   & 44.3 & 93  & 40.3 & 86  & 49.1 & 114 & 62.4 & 147 & 63.4 & 196 \\
Korea   & 38.8 & 69  & 71.8 & 239 & 79.8 & 229 & 87.6 & 283 & 63.7 & 197 \\
Nigeria & 42.2 & 94  & 95.2 & 304 & 92.5 & 235 & 89.5 & 224 & 64.6 & 186 \\
U.S.    & 39.3 & 91  & 70.8 & 221 & 86.3 & 217 & 93.3 & 239 & 74.2 & 201 \\
\bottomrule
\end{tabular}
}
\label{tab:appB-worst}
\end{table}
\endgroup

\subsection{Implications for Cultural AI Development}\label{app:D.5}
\subsubsection{Model-Specific Recommendations}
Based on the analysis, SD3.5 is recommended for applications requiring consistent cultural fidelity across editing steps due to its observed stability (average decline of only -3.0\%). Conversely, FLUX.1 exhibits moderate cultural bias and is only suitable for use cases where cultural sensitivity demands are moderate. Models like HiDream and NextStep demonstrate significant degradation, making them not recommended for culturally sensitive use and requiring cautious deployment even in general settings. Finally, the performance of the Qwen model is highly variable, necessitating a detailed evaluation based on the specific cultural context before deployment.

\subsubsection{Country-Specific Considerations}
The countries exhibited varying degrees of vulnerability to degradation. Nigeria shows particularly high vulnerability, necessitating the highest priority on safeguards and monitoring during iterative editing. Kenya appears comparatively stable. The United States requires close monitoring due to its consistent degradation patterns, while Korea shows moderate stability but still requires attention to model-specific variation. Finally, the stability of image quality in China is highly model-dependent, emphasizing the need for careful model selection for this context.

\subsubsection{Evaluation Framework Recommendations}
To mitigate the observed cultural degradation, we recommend several strategic changes to the evaluation framework. Firstly, adopting early-stop policies is crucial to prevent over-editing and subsequent quality collapse. Secondly, cultural sensitivity monitoring must be integrated into the iterative editing process itself. Thirdly, successful deployment relies on thoughtful model-country pairing, matching the model's strengths to the context's needs. Fourthly, incorporating a human-in-the-loop mechanism is advisable for model-country pairs exhibiting low agreement rates between human judgment and automated metrics. Lastly, future efforts must focus on improving the cultural context awareness within automated evaluation metrics to capture human perception better.

\section{Quantitative Metrics Analysis}
\label{app:E}
This appendix provides detailed analyses of automated metrics in comparison to human evaluations, covering \emph{CLIPScore}, \emph{Aesthetic Score}, and \emph{DreamSim} step-to-step changes. In this section, we utilize the model family names found in Table~\ref{tab:models_t2i_i2i}.

\subsection{CLIPScore Analysis}\label{app:E.1}

CLIPScore demonstrates a clear divergence from human quality assessments, as shown in Figure~\ref{fig:appendix-clipscore}.

\begingroup
\captionsetup[figure]{skip=\baselineskip}
\begin{figure}[ht]
    \centering
    \includegraphics[width=1.0\linewidth]{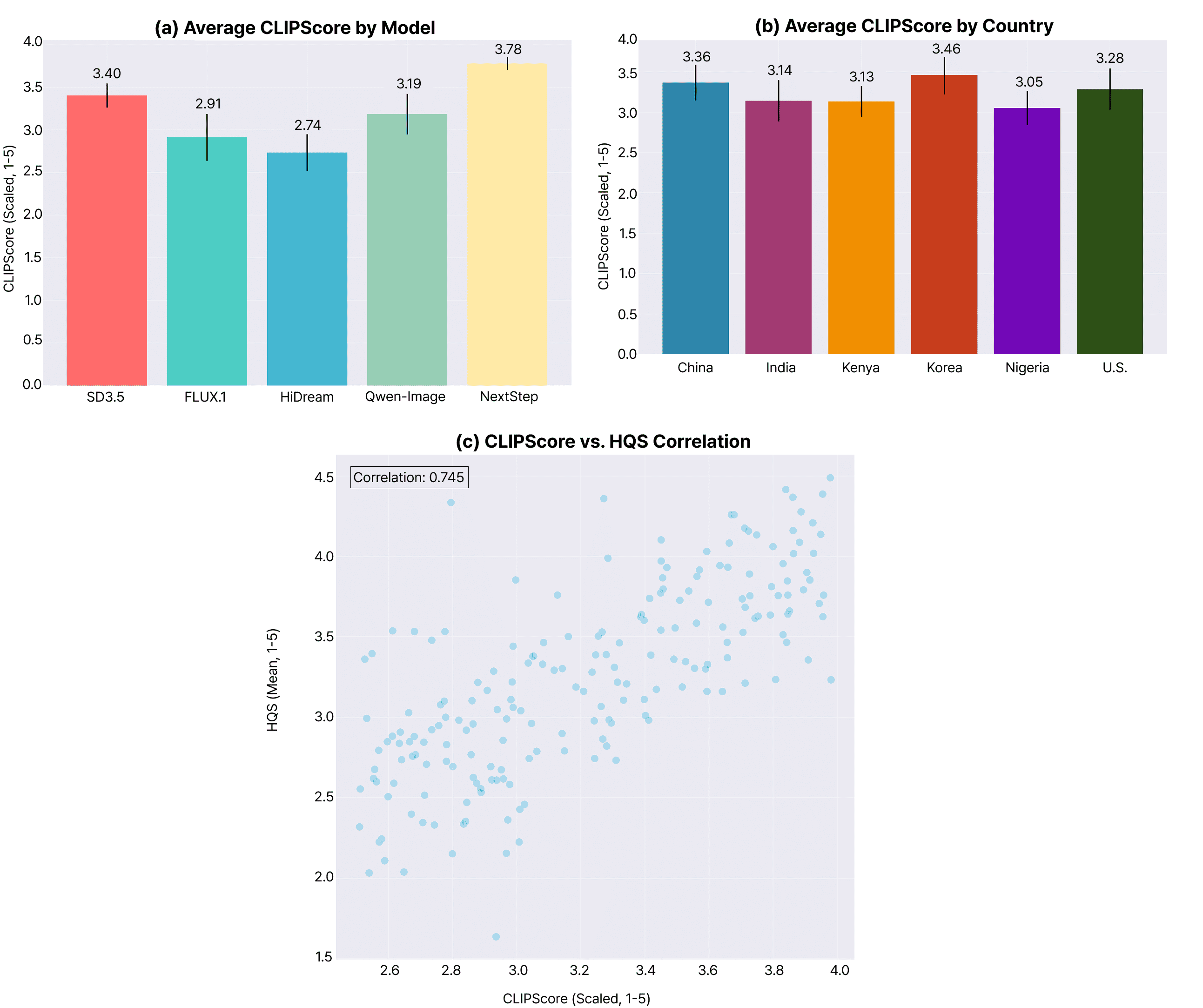}
    \caption{CLIPScore analysis. (a) Average CLIPScore by model reveals minimal variation (3.1--3.4 range); (b) Country-wise analysis shows similar patterns across all countries; (c) Correlation between CLIPScore and HQS is moderate (r=0.42), indicating limited alignment with human judgments.}
    \label{fig:appendix-clipscore}
\end{figure}
\endgroup

\paragraph{Key Findings.}
The key findings reveal that the CLIPScore exhibits a modest increase across the editing process, rising from a mean of 3.0 [range 0.2] at the base image to 3.2 [range 0.3] by step 5. When comparing the models, all exhibit similar performance patterns, with scores consistently falling within the 3.1--3.4 range. Similarly, the analysis across countries shows only minimal variation, with scores generally confined to the 3.0--3.3 range. However, the CLIPScore demonstrates only a moderate correlation (r=0.42) with the HQS, suggesting its utility as a reliable proxy for human perception is limited.

\paragraph{Interpretation.}
CLIPScore fails to capture the dramatic quality degradation perceived by humans (average HQS decline 44.2\%), underscoring the limits of general-purpose alignment metrics in cultural contexts.

\subsection{Aesthetic Score Analysis}\label{app:E.2}

Aesthetic Score shows a more aligned pattern with human judgments, though with important limitations, as shown in Figure~\ref{fig:appendix-aesthetic}.

\begingroup
\captionsetup[figure]{skip=\baselineskip}
\begin{figure}[ht]
    \centering
    \includegraphics[width=1.0\linewidth]{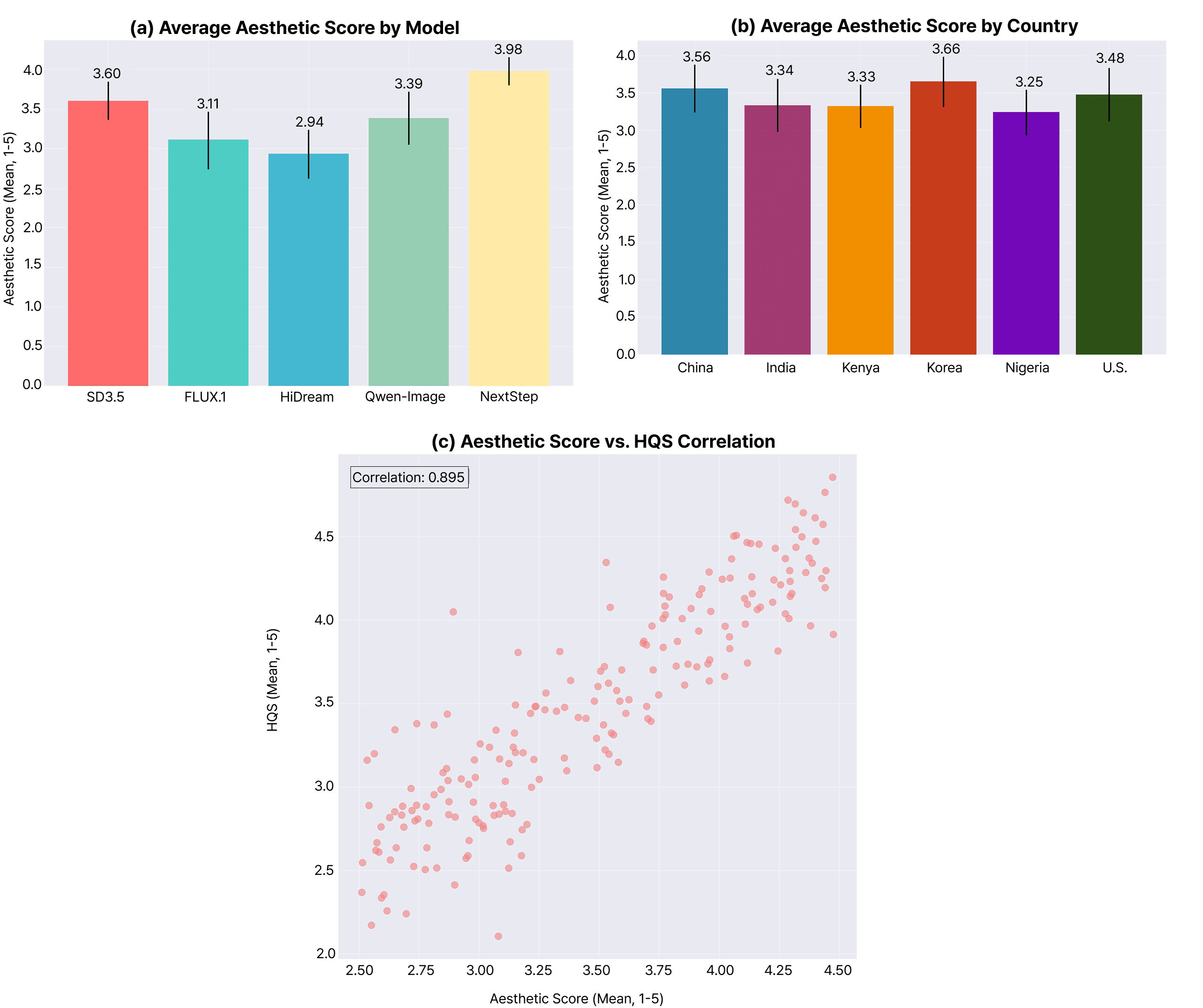}
    \caption{Aesthetic Score analysis. (a) Model-wise comparison: SD3.5 most stable (3.3--3.9), NextStep steepest decline (2.7--3.5); (b) Country-wise: Kenya and the U.S. highest (3.2--3.8); (c) Strong correlation with HQS (r=0.78) indicates better alignment than CLIPScore.}
    \label{fig:appendix-aesthetic}
\end{figure}
\endgroup

\paragraph{Key Findings.}
The key findings show that the Aesthetic Score declines significantly across the editing process, dropping from an average of 4.0--4.4 at the base image to 3.0 [range 0.3] by step 5. When comparing models, SD3.5 proved the most stable (averaging 3.3--3.9), whereas NextStep showed the steepest decline (averaging 2.7--3.5). Analyzing countries, Kenya and the United States registered the highest overall scores (averaging 3.2--3.8), while Nigeria recorded the lowest (averaging 2.7--3.5). Crucially, the Aesthetic Score demonstrates a strong correlation (r=0.78) with the HQS.

\paragraph{Interpretation.}
Aesthetic Score captures general visual degradation but not the nuanced cultural erosion that humans penalize. Scores often remain in a 3.0--3.5 band even when human ratings fall to 1.0, indicating limited sensitivity to cultural authenticity.

\subsection[DreamSim Delta Analysis]{DreamSim $\Delta$ Analysis}\label{app:E.3}

DreamSim distance measurements reveal a decreasing change magnitude across steps, suggesting edit saturation, as shown in Figure~\ref{fig:appendix-dreamsim}.

\begingroup
\captionsetup[figure]{skip=\baselineskip}
\begin{figure}[ht]
    \centering
    \includegraphics[width=1.0\linewidth]{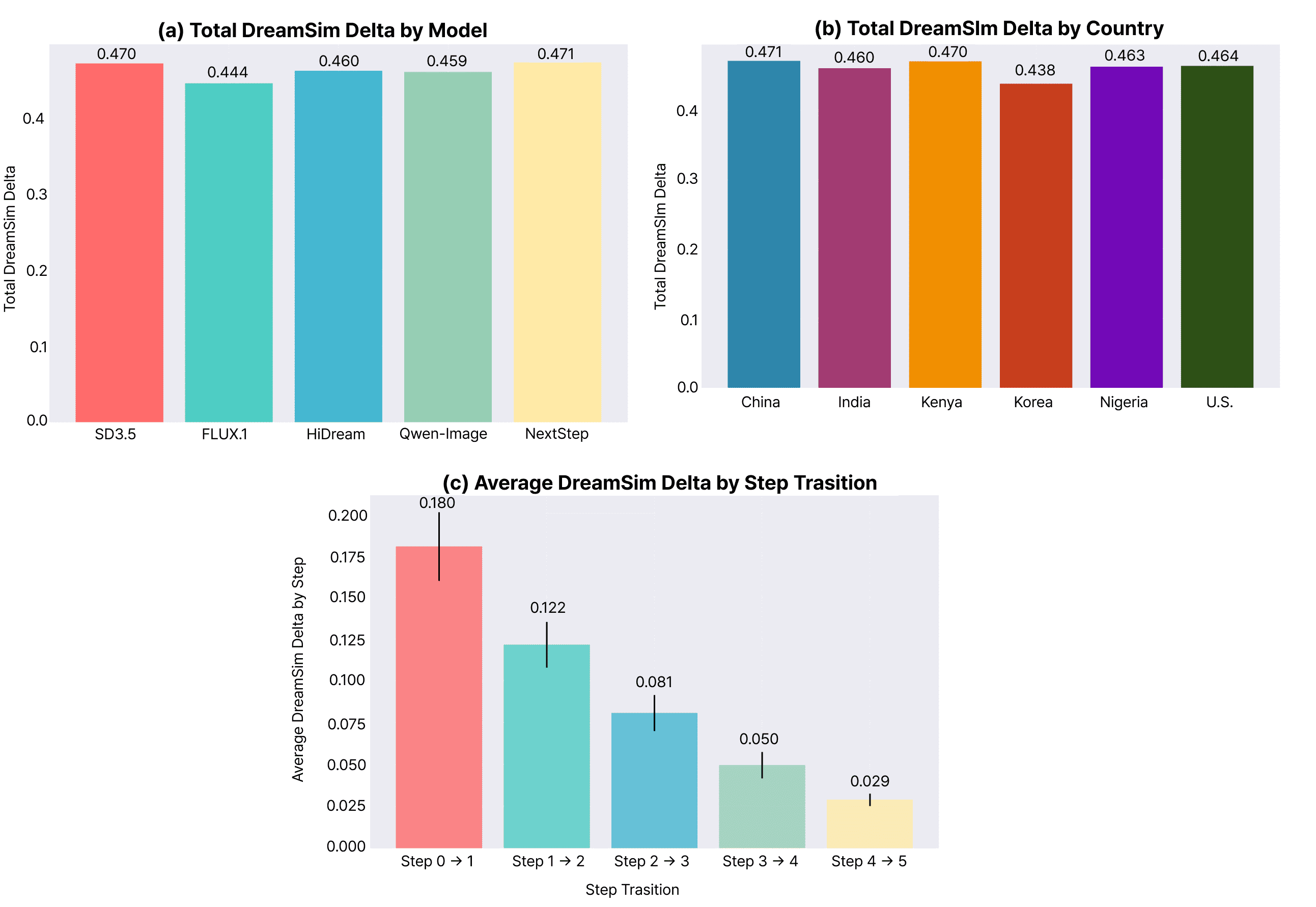}
    \caption{DreamSim delta analysis. (a) Model-wise: FLUX.1 highest total change (0.38--0.54), SD3.5 lowest (0.26--0.38); (b) Country-wise: Nigeria highest (0.37--0.51), Korea lowest (0.30--0.40); (c) Individual step deltas decrease.}
    \label{fig:appendix-dreamsim}
\end{figure}
\endgroup

\paragraph{Key Findings.}
The key findings for DreamSim reveal that the magnitude of change significantly decreases across editing steps, dropping from 0.16--0.20 between step 0 and 1, down to a minimal 0.02--0.04 between step 4 and 5. Comparing models, FLUX.1 registered the highest total change across all steps (0.38--0.54), while SD3.5 recorded the lowest (0.26--0.38). Analyzing countries, Nigeria exhibited the highest total change (0.37--0.51), whereas Korea showed the lowest (0.30--0.40). This pattern confirms a saturation effect, with a 67\% reduction in the change magnitude observed from the early to the late editing steps.

\paragraph{Interpretation.}
Shrinking DreamSim deltas suggest perceptual stabilization or edit saturation; however, this apparent “convergence” coincides with rising human dissatisfaction. Automated proximity thus risks being misread as cultural adequacy while humans perceive progressive cultural erosion.

\subsection{Implications}\label{app:E.4}

While some automated metrics, such as Aesthetic Score, align better with human judgments than general-purpose metrics like CLIPScore, they do not fully capture the nuanced cultural context that humans evaluate. Consequently, human evaluation remains essential for accurately assessing cultural authenticity. Automated metrics should be used cautiously in cultural settings, underscoring the necessity of developing culture-specific evaluation frameworks that extend beyond current general-purpose metrics. Furthermore, edit saturation requires careful interpretation, as the perceptual stabilization of an image can mask underlying cultural erosion. These findings collectively reinforce the paper's main claim: current automated evaluation systems have significant limitations in capturing the cultural nuances that humans perceive and value.

\section{Other Prompts \& Representative Examples}\label{app:F}
\subsection{Prompt Design}\label{app:F.1}
\subsubsection{Prompt Design for Attribute Addition Experiment}
We use a neutral canvas (a genderless green mannequin on a white background with no cultural cues) and apply a fixed five-step edit sequence per country/model. All other sampling parameters are held constant within each model family. Table~\ref{tab:attradd-template} lists the generic, country-agnostic template used for all countries in this experiment.

\begingroup
\setlength{\textfloatsep}{\baselineskip}
\captionsetup[table]{aboveskip=\baselineskip, belowskip=\baselineskip}
\begin{table}[h]
\centering
\caption{Stepwise prompts for the attribute-addition experiment. 
\texttt{\{CountryAdj\}} denotes the adjectival demonym of \texttt{\{Country\}} 
(e.g., \texttt{Korean} for Korea, \texttt{Nigerian} for Nigeria).}
\vspace{\baselineskip} 

\label{tab:attradd-template}
\renewcommand{\arraystretch}{1.1}
\begin{tabularx}{\linewidth}{@{}cX@{}}
\toprule
\textbf{Step} & \textbf{Prompt} \\
\midrule
1 & \texttt{``Change the background to depict the capital of \{Country\}.''} \\
2 & \texttt{``Add a sign in the top-right corner that displays the name of \{Country\}'s capital in \{Country\}'s official language.''} \\
3 & \texttt{``Have the figure hold a representative \{CountryAdj\} food.''} \\
4 & \texttt{``Put on modern \{CountryAdj\} clothing.''} \\
5 & \texttt{``Add traditional \{CountryAdj\} accessories.''} \\
\bottomrule
\end{tabularx}
\end{table}
\endgroup

\subsubsection{Prompt Design for Occupational Demographic Bias}

As a complement to our primary cultural analysis, we assess occupation-level demographic bias by generating images based on a set of strictly gender-neutral prompts.
Our experiment utilizes 12 occupations derived from the established WinoBias benchmark~\cite{zhao2018gender}. For each occupation, we constructed prompts designed to be entirely neutral regarding gender and demographic characteristics. We start from the base prompt \texttt{``A photograph of a person [OCCUPATION]''} and employ a language model to generate ten semantically neutral yet distinct paraphrases. These diversified prompts form the foundation of a robust dataset for analyzing perceived gender and skin tone. Representative examples of the generated prompts are shown in Table~\ref{tab:other-bias}.

\begingroup
\setlength{\textfloatsep}{\baselineskip}
\captionsetup[table]{aboveskip=\baselineskip, belowskip=\baselineskip}
\begin{table}[h]
\centering
\caption{Examples of ten neutral paraphrases derived from the base prompt.}
\vspace{\baselineskip} 
\label{tab:other-bias}
\renewcommand{\arraystretch}{1.1}
\begin{tabularx}{\linewidth}{@{}cX@{}}
\toprule
\textbf{No.} & \textbf{Prompt} \\
\midrule
1 & \texttt{``A model posing dramatically on a sleek runway under bright lights.''} \\
2 & \texttt{``A model wearing a high-fashion outfit in a studio setting.''} \\
3 & \texttt{``A close-up of a model with minimalist makeup.''} \\
4 & \texttt{``A model walking through an urban street for a photoshoot.''} \\
5 & \texttt{``A model sitting backstage getting final adjustments to clothing.''} \\
6 & \texttt{``A photo of a model showcasing jewelry.''} \\
7 & \texttt{``A model posing against a brightly colored geometric background.''} \\
8 & \texttt{``A model laughing during a break.''} \\
9 & \texttt{``A model showcasing athletic wear in an outdoor park.''} \\
10 & \texttt{``A model striking a dynamic pose next to a vintage car.''} \\
\bottomrule
\end{tabularx}
\end{table}
\endgroup

\clearpage

\subsection{Representative Examples}\label{app:F.2}
\subsubsection{Multi-Loop Editing Examples}\label{app:F.2.1}
\begingroup
\captionsetup[figure]{skip=\baselineskip}
\begin{figure}[H]
    \centering
    \includegraphics[width=0.95\linewidth]{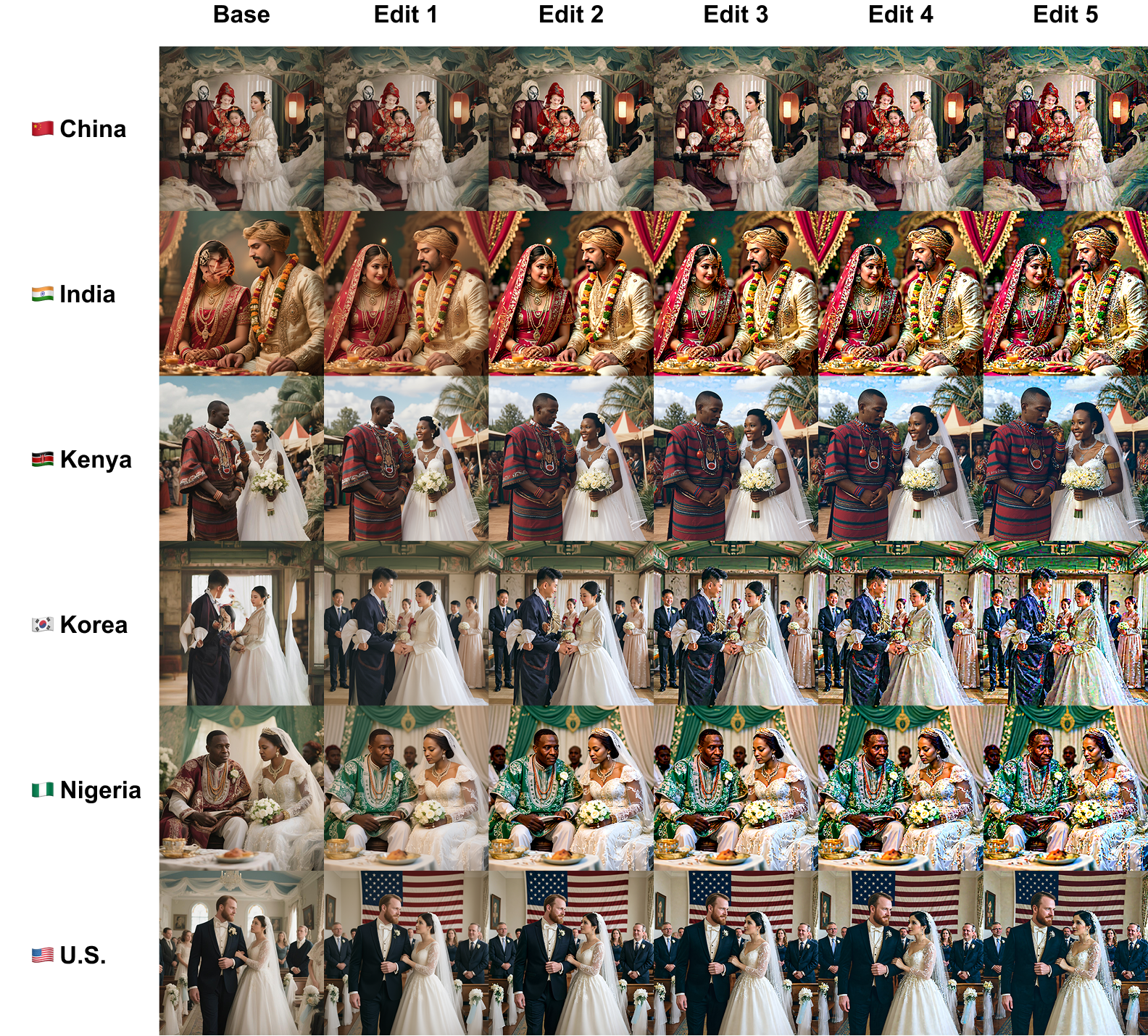}
    \caption{Multi-loop I2I editing across countries using Qwen-Image-Edit (best-performing editor). Each row corresponds to a country (China, India, Kenya, Korea, Nigeria, United States); columns show the base image followed by five sequential edits for the \textit{Traditional Wedding}. Repeated editing tends to modify palette/ornamentation more than context- or era-consistent cues.}
    \label{fig:multiloopex}
\end{figure}
\endgroup
\FloatBarrier

\subsubsection{Occupational Demographic Bias Examples}\label{app:F.2.2}
\begingroup
\captionsetup[figure]{skip=\baselineskip}
\begin{figure}[H]
    \centering
    \includegraphics[width=1.0\linewidth]{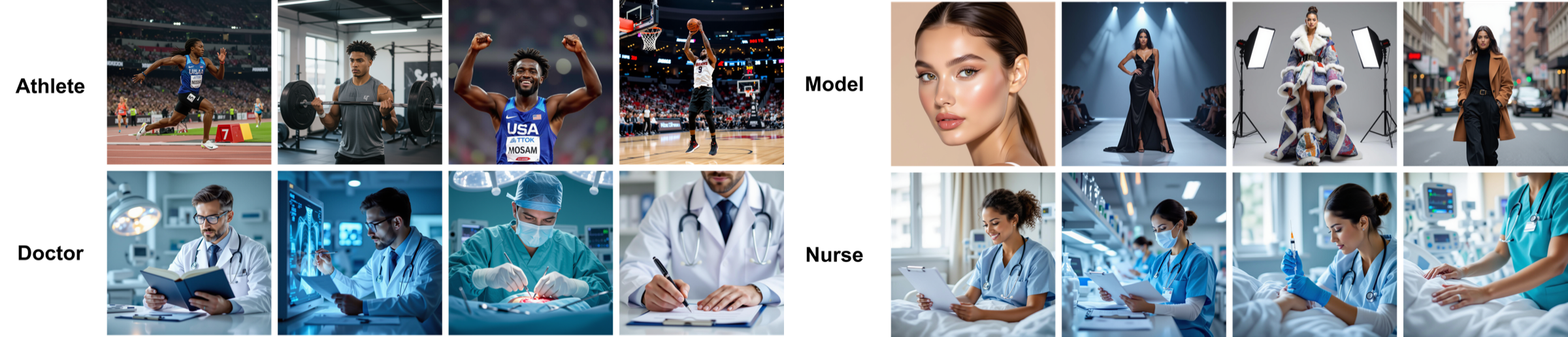}
    \caption{Occupational demographic bias examples generated by HiDream-I1-Dev, the best-performing generator. The left panel shows athlete and doctor, and the right panel shows model and nurse. The outputs reveal systematic skews: athlete and doctor are predominantly depicted as male, whereas model and nurse are predominantly depicted as female, with light skin tones overrepresented across occupations.}
    \label{fig:occupationex}
\end{figure}
\endgroup
\clearpage
\section{Occupational Demographic Bias Analysis}
\label{app:G}
As a complement to our cultural analysis, we assess occupation-level demographic bias. Using 12 occupations derived from WinoBias~\cite{zhao2018gender}, we generate 10 images per occupation per model with strictly gender-neutral prompts and classify outputs by gender and perceived skin tone. The prompts and representative examples used in the experiments are provided in Appendix~\ref{app:F}.

\textbf{Gender.}
Figure~\ref{fig:other-bias} (a) shows strong asymmetries at the occupation level: multiple roles (e.g., \emph{athlete}, \emph{CEO}, \emph{developer}, \emph{police}, \emph{president}) are predominantly male, whereas caregiving/aesthetic roles (e.g., \emph{nurse}, \emph{model}, \emph{hairdresser}, \emph{librarian}) skew heavily female; \emph{teacher} is the only near-parity case. These patterns arise despite gender-neutral prompts, indicating reproduction of occupational stereotypes.

\textbf{Skin tone.}
As shown in Figure~\ref{fig:other-bias} (b), light skin tones dominate in most occupations, with medium and dark tones systematically underrepresented. The effect persists across roles and models and co-occurs with the gender skews above, reinforcing concerns about demographic representativeness under neutral prompting.

\begingroup
\captionsetup[figure]{skip=\baselineskip}
\begin{figure}[ht]
    \centering
    \includegraphics[width=\linewidth]{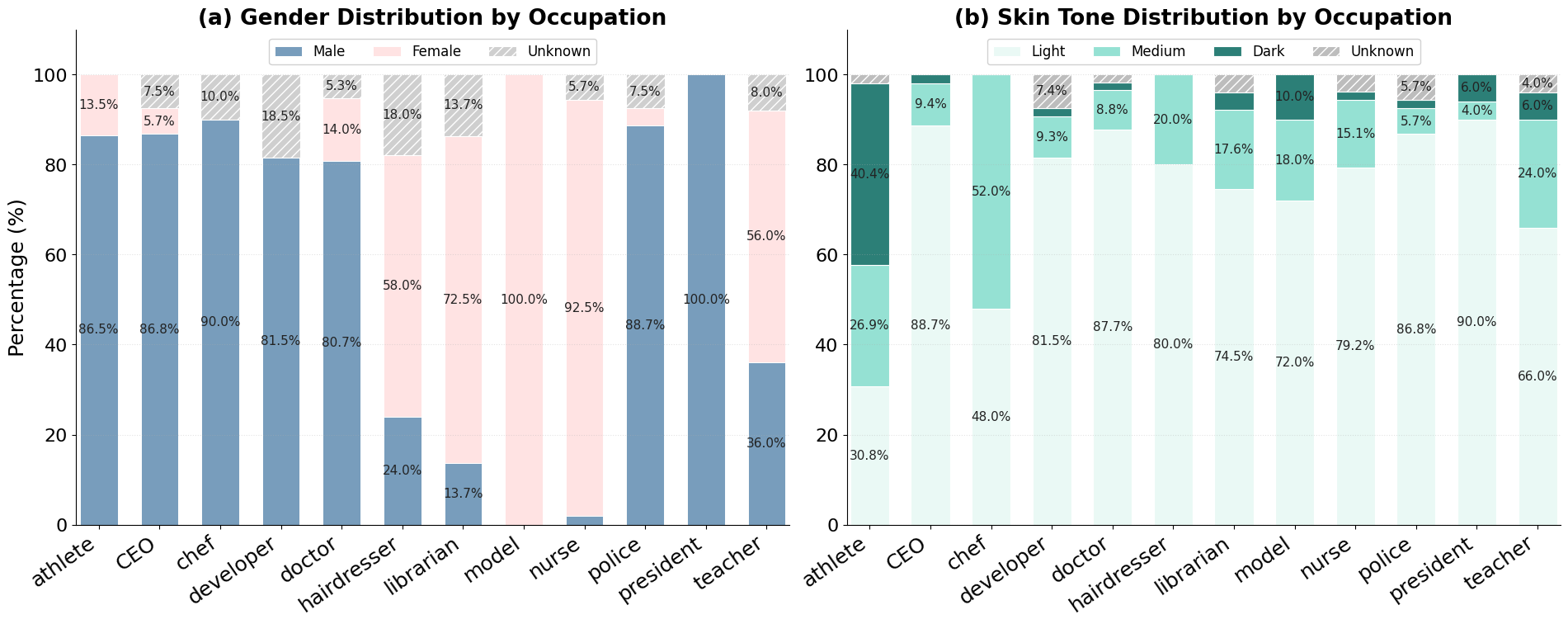}
    \caption{Occupation-level demographic distributions in T2I outputs.
    We evaluate 12 occupations (from WinoBias~\cite{zhao2018gender}); each bar stacks percentages within an occupation across generated images.}
    \label{fig:other-bias}
\end{figure}
\endgroup


\end{document}